\documentclass{article}

\PassOptionsToPackage{square, numbers, compress}{natbib}
\usepackage[final]{neurips_2022}

\usepackage[utf8]{inputenc} % allow utf-8 input
\usepackage[T1]{fontenc}    % use 8-bit T1 fonts
\usepackage{hyperref}       % hyperlinks
\usepackage{url}            % simple URL typesetting
\usepackage{booktabs}       % professional-quality tables
\usepackage{amsfonts}       % blackboard math symbols
\usepackage{nicefrac}       % compact symbols for 1/2, etc.
\usepackage{microtype}      % microtypography
\usepackage{xcolor}         % colors
\usepackage{graphicx}
\usepackage{subcaption}
\usepackage{multirow}
\usepackage{Definitions}
\usepackage{commands}
\usepackage{wrapfig}
\usepackage{hyperref}
\usepackage{xspace}
\usepackage[toc]{appendix}

\title{MGNNI: Multiscale Graph Neural Networks with Implicit Layers}

\author{
Juncheng Liu \ \ \ \ Bryan Hooi  \ \ \ \ Kenji Kawaguchi \ \ \ \  Xiaokui Xiao \\ 
National University of Singapore \\ 
\texttt{\{juncheng,bhooi,kenji,xiaoxk\}@comp.nus.edu.sg}
}
\begin{document}

\maketitle

\begin{abstract}
Recently, implicit graph neural networks (GNNs) have been proposed to capture long-range dependencies in underlying graphs.
In this paper, we introduce and justify two weaknesses of implicit GNNs: the constrained expressiveness due to their limited effective range for capturing long-range dependencies, and their lack of ability to capture multiscale information on graphs at multiple resolutions.
% However, there are two weaknesses of implicit GNNs: their constrained expressiveness due to their limited effective range for capturing long-range dependencies, and their lack of the ability to capture multiscale information on graphs at multiple resolutions.
To show the limited effective range of previous implicit GNNs, We first provide a theoretical analysis and point out the intrinsic relationship between the effective range and the convergence of iterative equations used in these models.
To mitigate the mentioned weaknesses, we propose a multiscale graph neural network with implicit layers (\ourmodel) which is able to model multiscale structures on graphs and has an expanded effective range for capturing long-range dependencies.
We conduct comprehensive experiments for both node classification and graph classification to show that \ourmodel outperforms representative baselines and has a better ability for multiscale modeling and capturing of long-range dependencies.
\end{abstract}
\section{Introduction}
% background intro for GNNs and why is important.
In recent years, graph neural networks (GNNs) have been widely adopted on graph-related tasks, such as node classification, link prediction, and graph classification \citep{wu2020comprehensive}.
In general, GNNs utilize both node attributes and graph topology to produce meaningful node representations for downstream applications. 
To achieve this, most modern GNNs follow a ``message passing'' mechanism: at each iteration, they iteratively aggregate representations of neighboring nodes of each node with its own representation to generate new representations. During this process, each iteration is typically parameterized as a single-layer neural network with learnable weights.
Many GNN models have been proposed by adopting different aggregations techniques (e.g., GCN with renormalization \citep{semi_GCN}, GAT with attentive aggregations on neighbors \citep{GAT}, and SGC \citep{SGC} using aggregations without non-linear activation).
% limitations of finite-gnns and introduce implicit gnns. How Implicit gnns relates to long-range dependencies. 
In spite of the effectiveness achieved by the aforementioned GNNs on different tasks, they fail to effectively capture long-range information on graphs, since a GNN with \textit{T} layers can only capture information up to \textit{T} hops away. 
% To explain, with predefined \textit{T} aggregation layers, these models can only capture the information up to \textit{T}-hops away.
% Therefore, for any given node, it always fails to capture dependencies with a range longer than \textit{T}-hops away. 

% Explain here why long-range dependencies are important. 
To overcome this deficiency of previous GNNs, recent work has proposed implicit graph neural networks \citep{IGNN,EIGNN,CGS} to effectively capture long-range dependencies. 
These implicit graph neural networks generally define a fixed-point equation as an implicit layer for aggregation and generate the equilibrium $Z^*$ as the node representations. 
To get the equilibrium, they either use an iterative solver to solve the equation or directly obtain a closed-form solution with guaranteed convergence. Meanwhile, they utilize implicit differentiation to achieve $\mathcal{O}(1)$ memory complexity when computing the gradients during the iterations. 
As mentioned in \citet{IGNN, EIGNN}, these models can be treated as a graph neural network with infinite layers which has the same transformation and shared weights in each layer.  
This makes them able to effectively capture long-range dependencies without excessive memory requirements as compared with previous GNNs. 
% discuss the limitation of previous implicit graph models (e.g., (1) not actually infinite depth, the effective range is limited; (2) they cannot capture multiscale information.)

Despite the superiority of implicit GNNs shown in several applications requiring long-range information, an important question --- what the farthest range these models can capture information from --- has not been studied. 
Although these models are usually claimed as GNNs with infinite depth \citep{IGNN, EIGNN}, in this paper, we first point out the \emph{effective range} of these models (i.e., the maximum hops they can effectively capture dependencies for each node) is actually bounded by a certain value. 
We provide analyses on the intrinsic relationship between the effective range and the convergence of the iterative equation used in implicit GNNs. 
% related to convergence speed?  
% slightly go to the details about this?
Specifically, these models usually use a contraction factor $\gamma$ to ensure the convergence of the iterative map \citep{EIGNN, CGS}, which indeed exponentially decays the distant information during the aggregation at the same time. 
This design inherently limits the effective range of propagation and hinders their ability to capture long-range dependencies. 

Besides the limited effective range, implicit GNNs also cannot effectively capture multiscale information on graphs, i.e., graph features at various scales. 
In contrast, several GNNs without implicit layers \citep{JKNet, NGCN,mixhop, GCNII} have been proposed to utilize multiscale information to improve the model capacity.
% and they demonstrate the effectiveness of mixing graph information from neighbors at various distances. 
For example, \citet{JKNet} proposes JKNet which leverages different neighborhood ranges by skip connections and adaptive aggregations of hidden representations at different layers. 
MixHop \citep{mixhop} learns neighborhood mixing relationships by mixing hidden representations at various distances. 
These explicit GNN models demonstrate the effectiveness of utilizing multiscale information from neighbors at various scales. 
However, it is still not clear how to utilize multiscale information with implicit GNNs since there are no different ``layers'' in implicit GNNs that can be used for capturing multiscale information at different scales. 

Motivated by the above limitations of previous implicit GNNs, we propose our multiscale graph neural network with implicit layers (\ourmodel) which brings multiscale modeling into implicit GNNs and expands their effective range for capturing long-range dependencies.  We summarize the contributions of this work as follows: 
\setlist{leftmargin=*}
\begin{itemize}
	\item  We introduce the concept of effective range for implicit GNNs, and provide theoretical analyses on the effective range that previous implicit GNNs can capture distant information from. We then point out that their effective range is limited although previous models are generally regarded as GNNs with infinite layers. 
% 	We first provide theoretical analyses on the effective range that previous implicit GNNs can capture distant information from and point out the effective range is limited although previous models are generally regarded as GNNs with infinite layers. 
	\item  We propose \ourmodel as a new implicit GNN model with multiscale propagation to expand the effective range and capture underlying graph information at various scales.
	\item We conduct comprehensive experiments with synthetic datasets and real-world datasets on both node classification and graph classification to demonstrate that \ourmodel has better performance and a better ability to capture both long-range and multiscale information compared with other baselines. 
\end{itemize}

% \paragraph{Paper outline}

\section{Related work}
\paragraph{Implicit Models}
Implicit neural networks use implicit hidden layers which are \textit{implicitly} defined: the outputs are determined by the solutions of some underlying equations.
A notable advantage of these implicit models is that they can generally backpropagate through the fixed-point solution using implicit differentiation to achieve \textit{constant} memory complexity regardless of the ``depth'' of the network. 
There is an emerging interest in implicit layers in recent years \citep{DEQ, amos2017optnet, MDEQ, geng2021on}. 
To name a few, \citet{DEQ} propose the deep equilibrium model (DEQ) demonstrating the ability of implicit models in sequence modeling; Multiscale DEQ (MDEQ) \citep{MDEQ} brings multiscale modeling into implicit deep networks for image classification and semantic segmentation. 
\citet{kawaguchi2021theory} analyses the global convergence of deep linear implicit models and \citet{geng2021on} provide a gradient estimate for implicit models to avoid the costly exact gradient computation. 
\paragraph{Graph Neural Networks}
% background 
% explicit gnns with multiscale modeling 
% implicit gnns 
GNNs have been widely used in different tasks for graph-structured data. 
Even with different aggregation schemes (e.g., skip connection \citep{JKNet, GCNII} and attention \citep{GAT}), convolutional GNNs \citep{semi_GCN, SGC,JKNet} generally involve finite aggregation layers (usually less than 20 layers) with different learnable weights, which makes them unable to effectively capture long-range dependencies. 
Although RevGNN \citep{li2021training} is proposed with 1000 layers, it has to use deep reversible architectures \citep{gomez2017reversible}, which requires excessive amount of time for training. 
Inspired by implicit models \citep{DEQ, MDEQ} on image and text data, implicit graph neural networks \citep{IGNN, EIGNN, CGS} have been proposed to capture long-range information with constant memory complexity. 
Implicit GNNs generally define an aggregation equation and obtain the fixed-point solution of the equation as the outputs.
% , which potentially allows numerous aggregation steps until convergences. 
In particular, \citet{IGNN} propose IGNN where they ensure the well-poseness and use iterative solvers to obtain fixed-point solutions. \citet{EIGNN} propose EIGNN as a linear implicit GNNs where a closed-form solution is derived. \citet{CGS} construct an input-dependent linear iterative map for predicting the properties of a graph system. 
However, these implicit GNNs cannot model multiscale information in underlying graphs. In contrast, several explicit GNNs, such as JKNet \citep{JKNet}, MixHop \citep{mixhop}, and N-GCN \citep{NGCN}, have shown that multiscale information is helpful to improve the model capability. 
To fill the gap, our model \ourmodel brings multiscale modeling to implicit GNNs. 

\section{Preliminaries}
A graph is represented as $\mathcal{G}=(\mathcal{V}, \mathcal{E})$ which contains the node set $\mathcal{V}$ with $n$ nodes and the edge set $\mathcal{E}$. 
In practice, graph neural networks take the adjacency matrix $A \in \mathbb{R}^{n \times n}$ and the node feature matrix $X \in \mathbb{R}^{m \times n}$ of $\mathcal{G}$ as input data. 
For simplicity, considering unweighted adjacency matrix $A$, then $A_{i,j} = 1$ if $(i,j) \in \mathcal{E}$, for any two nodes $i, j \in \mathcal{V}$; otherwise $A_{i,j} = 0$.
Given input graph data $(\mathcal{G}, X)$, depending on different classification tasks, graph neural networks are required to provide a prediction $\hat{y}$ for a node or a graph to match the true label $y$.

% Briefly mention about the general graph neural network framework (i.e., message passing)
\paragraph{Aggregations in GNNs} 
GNNs typically employ a trainable aggregation process that iteratively pass the information from each node to its adjacent nodes, followed by a non-linear activation. Without loss of generality, a typical aggregation step at layer $l$ can be written as follows:
\begin{equation}
\label{eq:graph_aggregation}
Z^{(l+1)} = \phi (W^{(l)}Z^{(l)}S+ \Omega^{(l)}X),   
\end{equation}
where $Z^{(l)} \in \mathbb{R}^{h_{l} \times n}$ is the hidden states in the layer $l$ which stacks the state vectors of every nodes denoted as $z^{(l)} \in \mathbb{R}^{h_{l}}$; $S \in \mathbb{R}^{n \times n}$ is the normalized adjacency matrix; $W^{(l)} \in \mathbb{R}^{h_{l+1}\times h_l}$ and $\Omega^{(l)} \in \mathbb{R}^{h_{l+1} \times m}$ are the matrices of trainable weight parameters; $\phi$ denotes a non-linear activation function.  
Recently proposed GNN models use different forms of this graph aggregation process. For example, simplified graph convolution \citep{SGC} removes the non-linear activation, use only one weight matrix $W$, and sets $Z^{(0)} = X$ and $\Omega = 0$.  

In addition to GNNs with explicitly defined layers, GNNs with implicit layers \citep{IGNN, EIGNN, CGS} also follow a similar aggregation form, but with tied weight matrices $W$ and $\Omega$ at each iteration step.
% For these implicit GNN models, the aggregation step is generally changed to $Z^{(l+1)} = \phi (WZ^{(l)}S+ \Omega X)$ with some constraints to impose the convergence of this iterative mapping.
For these implicit GNNs, the aggregation step is generally changed to $Z^{(l+1)} = \phi (WZ^{(l)}S+ \Omega X)$. 
Given such an aggregation step, implicit GNNs can be seen as iterating the aggregation step an infinite number of times until convergences. 
To ensure the convergence, IGNN \citep{IGNN} enforces $\|W\|_{\infty} \leq \kappa / \lambda_{p f}(A)$ with $\kappa \in [0, 1)$, where $\lambda_{pf}$ is the Perron-Frobeius (PF) eigenvalue \citep{berman1994nonnegative}. 
% Through another way, EIGNN \citep{EIGNN} and CGS \citep{CGS} introduce a contraction factor $\gamma$ in the aggregation step. 
EIGNN \citep{EIGNN} and CGS \citep{CGS} instead achieve this by introducing a contraction factor $\gamma$ in the aggregation step. 

Using EIGNN as an example, it defines the aggregation as an iterative mapping with a contraction factor $\gamma$ as follows: 
\begin{equation}
\label{eq:Implicit_propagation}
    Z^{(l+1)} = \gamma g(F)Z^{(l)}S + X,   
\end{equation}
where $\gamma \in [0, 1)$ and $g(W)$ is a bounded mapping that projects the trainable weight matrix $F$ into a Frobenius norm ball of radius < 1.
Given the iterative mapping, implicit GNNs obtain the equilibrium states $Z^{*} = \phi (W Z^{*} S+ \Omega X)$ as the final representations by using either root-finding approaches or closed-form solutions.

% Then mention how this framework includes GCN, an other implicit graph neural network.
% \TODO{Discuss how conventional GNNs are usually different form of this aggregation. Then implicit graph neural networks fall in the similar form but different manner with shared weights}

% \TODO{Take EIGNN as an example. Show the equation with the damping factor. Mention the condition of $\gamma$ for convergence. Then go to Sec 4 for analyzing the effective range.}
\section{Effective range of previous implicit GNNs}
\label{sec:range_analysis}
Implicit deep learning is considered as a method to increase the effective depth of deep neural networks \citep{DEQ, ramzi2022shine}. 
Previous works on implicit GNNs (including IGNN \citep{IGNN} and EIGNN \citep{EIGNN}) claim that these models can be viewed as an infinite-layer GNN. However, in this section, we point out that the effective range within which previous implicit GNNs can capture the long-range information is actually bounded. In other words, their abilities for capturing long-range dependencies are still restricted. 
We provide analyses revealing the intrinsic relationship between the effective range and the convergence of the iterative equation in implicit GNNs.
In addition, we also provide empirical results to support our analyses. 

To investigate the effective range, we first provide an analysis on sensitivity, i.e., how changes in node features of node $p$ affect the equilibrium of a distant node $q$. 
\begin{theorem}
\label{decaying sensitivity} 
Given two nodes $p$ and $q$ that are $h$-hops apart, using Equation \eqref{eq:Implicit_propagation} for propagation, if we perturb node features $X_{:,p}$ of node $p$ by $\Delta X_{:,p} \in \mathbb{R}^{m}$, the L2 norm of the change in node $q$'s equilibrium $\|\Delta Z^{*}_{:, q}\|$ is upper bounded as follows: 
\begin{equation}
\label{eq: sensitivity}
%   \Delta Z^{*}_{:, q} < (h_{\theta} -h)\gamma^{h}g^{h}(F)(\Delta X_{:,p}S^{h}_{p,q}), 
%   \Delta Z^{*}_{:, q} \leq \frac{\epsilon}{1-\gamma}g^{h}(F)(\Delta X_{:,p}S^{h}_{p,q}), 
   \|\Delta Z^{*}_{:, q}\| \leq \frac{\gamma^h}{1-\gamma} \|g^{h}(F)\Delta X_{:,p}S^{h}_{p,q}\|. 
\end{equation}
% assuming $\gamma^i < \epsilon$ for all $i > h_{\theta}$ with $h_{\theta} > h$.
% assuming $\gamma^i \leq \epsilon$ for all $i \geq h$.
\end{theorem}
% \TODO{briefly explain the proof sketch.} 
The complete proof can be found in Appendix \ref{proof of decaying sensitivity}.
% To show a new bound given gamma < epsilon, and then explain why it would be limited range. That is the amount of change on equilibrium would go down below the numerical limit. Then it cannot be handled by computers
% In Sec 5, we show that we have larger changes on the equilibrium.

At first glance, the norm of the change in equilibrium $\|\Delta Z^{*}_{:,q}\|$ would not be zero no matter how large the distance between node $p$ and $q$. 
However, Theorem \ref{decaying sensitivity} shows that $\|\Delta Z^{*}_{:,q}\|$ decays exponentially with distance along the graph.
Therefore, in practice, this change will fairly quickly fall below the roundoff error in floating-point numbers or the stopping criterion in the iterative solver used to obtain the fixed-point solution, then the change $\Delta X_{:,p}$ on node $p$ cannot affect the equilibrium of node $q$. 
% However, in practice, if the change $\Delta Z^{*}_{:,q}$ goes below the roundoff error in floating-point numbers or a stopping criterion in the iterative method, then the change $\Delta X_{:,p}$ on node $p$ cannot affect the equilibrium of node $q$. 
% This is what we call the effective range.
This is the intuition about what we call the effective range. To formalize it, we provide the definition of $\theta$-effective range as follows:
\begin{definition}
\label{def: effective range}
For any given error parameter $\theta > 0$, the $\theta$-effective range $h$ is the maximum integer such that exists some pairs of nodes $p$ and $q$ that are $h$-hop apart, when node features $X_{:,p}$ of node $p$ are perturbed by $\Delta X_{:, p}$, the L2 norm of the change in node $q$’s equilibrium $\left\lVert \Delta Z^{*}_{:,q}\right\rVert > \theta $.
\end{definition}
By Definition \ref{def: effective range}, we know that, given any $h' > h$, for \textbf{all pairs} of node $p$ and $q$ that are $h'$-hop apart, the L2 norm of the change $\left\lVert \Delta Z^{*}_{:,q}\right\rVert \leq \theta $, which means the equilibrium of node $q$ cannot be affected in practice.  
To analyse the $\theta$-effective range, we can derive the corollary of Theorem \ref{decaying sensitivity}:
\begin{corollary}
\label{corollary: limited effective range}
With Equation \eqref{eq:Implicit_propagation} for propagation, given any error constant $\theta > 0$, the $\theta$-effective range $h$ is upper bounded:
$h < \frac{\ln (\theta(1-\gamma))}{\ln \gamma}$. 
Therefore, if node features
$X_{:,p}$ of node $p$ are perturbed, the perturbation can only affect the equilibrium of nodes which are up
to $\frac{\ln (\theta(1-\gamma))}{\ln \gamma}$-hop away from $p$.
\end{corollary}

% To analyse the effective range, we derive the corollary of Theorem \ref{decaying sensitivity}:
% \begin{corollary}
% \label{corollary: limited effective range}
% With Equation \eqref{eq:Implicit_propagation} for propagation, given any small error constant $\theta$, if node features $X_{:,p}$ of node $p$ are perturbed, the perturbation can only affect the equilibrium of nodes which are up to $\frac{\ln (\theta(1-\gamma))}{\ln \gamma}$-hop away from $p$. 
% \end{corollary}
The complete proof can be found in Appendix \ref{proof of limited effective range}.
% \TODO{To directly point out the effective range is $h_{\theta}$, then after this range, no information can be received.}
% After quantifying the influence on the equilibrium, we can derive the effective range within which the perturbation on a node can be captured by another distant node. 
% \begin{corollary}
% With Equation \eqref{eq:Implicit_propagation} for propagation, assuming $\gamma^i \approx 0$ for all $i > h_{\theta}$, if node features $X_{:,p}$ of node $p$ are perturbed, the perturbation can only affect the equilibrium of nodes which are up to $h_{\theta}$-hop away from $p$. 
% \end{corollary}
% \begin{proof}
% This is the corollary of Theorem \ref{decaying sensitivity}. The change on node $q$'s equilibrium $\Delta Z^{*}_{:,q} = 0$ if the distance between node $q$ and $p$ is larger than $h_{\theta}$.
% \end{proof}
Note that the above analysis is directly applicable to two recent Implicit GNN models with a contraction factor $\gamma$, i.e., EIGNN \citep{EIGNN} and CGS \citep{CGS}.

\begin{wrapfigure}{r}{0.4\textwidth}
%   \vspace{-1mm}
  \begin{center}
    \includegraphics[width=0.39\textwidth]{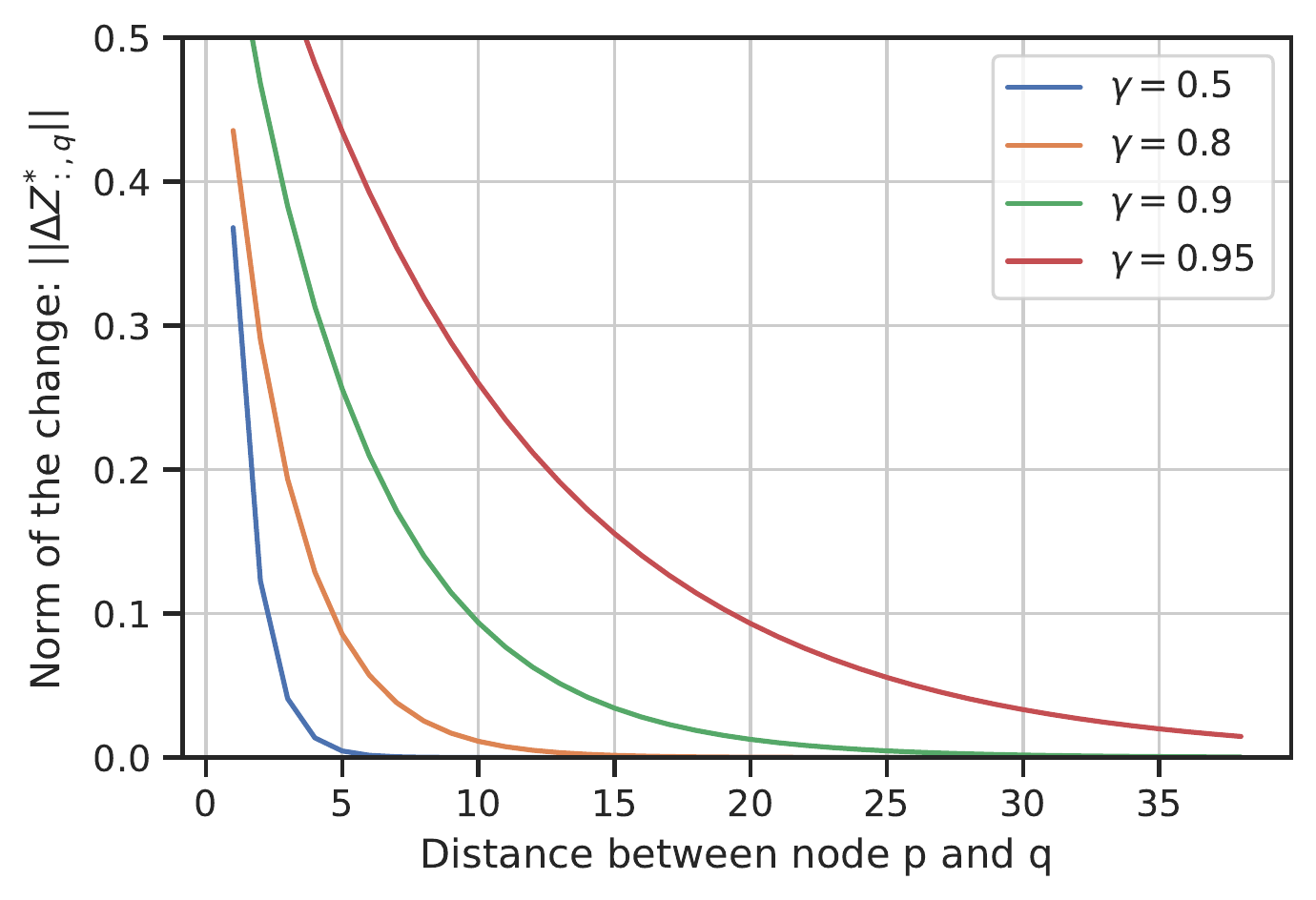}
  \end{center}
  \caption{The norm of the change of equilibrium $\|\Delta Z^{*}_{:,q}\|$ decays as node $q$ becomes further from $p$.}
  \label{fig:varing_gamma}
  \vspace{-2mm}
\end{wrapfigure}

% \TODO{Empirical results to support the analyses} 
Asides from the above analysis, we further verify the theoretical analysis with synthetic experiments on the same chain dataset as in EIGNN \citep{EIGNN} and IGNN \citep{IGNN}, where the task requires simply passing information from one end to the other of a chain graph. 
See Appendix \ref{appendix:synthetic_exp} for detailed settings. 
% Basically, the label information is encoded in the node features of the first node on the chain, while the node features of all other nodes are all zeros.
We use the basic form of EIGNN \citep{EIGNN} with iterative methods (i.e., iterating Equation \eqref{eq:Implicit_propagation} to find the equilibrium) as the model.
% The node features are all zeros at the beginning. 
% We consider the first node as node $p$ and perturb the node features of node $p$ by adding ones.
We follow the same experimental setup in EIGNN and then perturb node features by masking the features of the starting node $p$ to all zeros. 
To support our theoretical analysis, we investigate how the change of node $q$'s equilibrium (i.e., $\Delta Z^{*}_{:,q}$) behaves as $q$ gets farther away from $p$. 
% To support our analysis, we consider the first node as node $p$ and observe how the equilibrium of node $q$ changes as $q$ are farther away from $p$.
In Figure \ref{fig:varing_gamma}, we show that $\|\Delta Z^{*}_{:, q}\|$ decays as node $q$ gets further from $p$ in terms of the distance.
For example, with $\gamma=0.5$, $\|\Delta Z^{*}_{:, q}\|$ becomes numerically 0 when $q$ and $p$ are around 25 hops apart, which indicates that node $q$ can no longer receive any information from node $p$ at this distance. 
% Eventually, the norm of $Z^{*}_{:, q}$ becomes numerically 0, which indicates that node $q$ can no longer receive any information from node $p$ at this distance. 
% \TODO{Maybe show a similar figure of IGNN in appendix and slightly mention here. (Although our analysis is based on EIGNN, we can use empirical evidences to show that this phenomenon is common in implicit GNNs not only in EIGNN?)} 

% \paragraph{Discussion on the model expressiveness} 
The above theoretical analysis and empirical evidences show how different values of $\gamma$ can constrain the expressiveness of implicit GNNs, i.e., the range within which they can effective capture long-range dependencies. 
% the ability of capturing long-range dependencies to how far. 
In short, smaller $\gamma$ results in a shorter effective range of implicit GNNs. 
A straightforward strategy to expand the effective range is to increase $\gamma$ (e.g., to close to 1).
However, a large $\gamma$ can cause instability and difficulty for the convergence of iterative mapping, which would empirically compromise the efficiency of iterative solvers as found in our experiments.
% (\TODO{Need to provide evidences for this point in the appendix?}). 
We provides the empirical evidences for this in Appendix \ref{appendix: inefficiency and instability}.
% The constrained expressiveness of implicit GNNs is not surprising since the iterative mapping need to be contractive in the view of the Banach fixed-point theorem.
% Regarding the constrained expressiveness of implicit GNNs, it might not be surprising since the iterative mapping need to be contractive from the view of Banach fixed-point theorem \citep{banach1922operations}.
This raises the question: \textit{how can we capture longer range dependencies while ensuring the convergence of the iterative mapping?} 
Our multiscale \ourmodel approach aims to answer this question.

\section{Multiscale implicit graph neural networks}
Previous implicit GNNs \citep{IGNN, EIGNN, CGS} propagate the information from 1-hop neighbors while applying a decay with the contraction factor $\gamma$ in each iterative step (e.g., using Equation \eqref{eq:Implicit_propagation}.
As analysed in Section \ref{sec:range_analysis}, this design inherently limits the effective range of propagation since the information decays exponentially as the range grows linearly.
Besides the limited effective range, previous implicit GNNs also cannot capture multiscale information on graphs similarly to explicit GNNs, such as JKNet \citep{JKNet} and MixHop \citep{mixhop}, which combine information at different scales of the graph.

Motivated by these limitations, we propose  multiscale graph neural networks with implicit layers (\ourmodel) which can first expand the effective range to capture long-range dependencies and then capture information from neighbors at various distances. 
% In \ourmodel, we defines multiple propagation components with different scales. 
\ourmodel contains multiple propagation components with different scales and learns a trainable aggregation mechanism for mixing latent information at various scales. 
\subsection{The \ourmodel model}
A single $m$-scale propagation module in \ourmodel model is defined as the following iterative mapping: 
\begin{equation}
\label{eq:multiscale_prop}
    Z^{(l+1)} = \gamma g(F)Z^{(l)}S^{m} + f(X, \mathcal{G}), 
    % \lim_{l\rightarrow\infty} Z_i^{(l+1)} = Z^{*} s.t. Z_i^{*} &= \gamma g(F)Z_{i}^{*}S^{m_i} + f(X, \mathcal{G})
\end{equation}
where $\gamma \in [0, 1)$ and $m$ denotes a hyperparameter for the graph scale (i.e., the power of adjacency matrix). $f(X, \mathcal{G})$ is a parameterized transformation on input features and graphs, and $g(F)$ is normalized weight matrix defined as: 
\begin{equation}
    g(F)=\frac{1}{\left\|F^{\top} F\right\|_{\mathrm{F}}+\epsilon_{F}} F^{\top} F
\end{equation}
with an arbitrary small $\epsilon_F > 0 $. 
Note that in multiscale propagation, at each iterative step, the model can capture the information along with a m-step path, while previous implicit GNNs only consider 1-hop neighbors. 
In this way, \ourmodel is able to capture dependencies within a longer range over iterations. 
Now we provide the analysis to show that the iterative mapping in \ourmodel (i.e., Equation \eqref{eq:multiscale_prop}) converges to a unique equilibrium $Z^{*}$: 
\begin{equation}
    \lim_{l\rightarrow\infty}Z^{(h)}= Z^{*} \quad \text{s.t.} \quad Z^{*} = \gamma g(F)Z^{*}S^{m} + f(X, \mathcal{G}).
\end{equation}
\begin{theorem}
\label{theorem:convergence}
Given the bounded damping factor $\gamma \in [0, 1)$, the proposed iterative map for propagation (i.e., Equation \eqref{eq:multiscale_prop}) is a contraction mapping and the unique fixed-point solution $Z^{*}$ can be obtained by iterating Equation \eqref{eq:multiscale_prop}.  
\end{theorem}
This can be proved by using the properties of matrix vectorization and the Kronecker product with the Banach fixed Point Theorem. The complete proof is given in Appendiex \ref{proof: convergence}. 
% \begin{proof}
% (Sketch) We use the property of the matrix vectorization and the Kronecker product to show the contraction mapping, and use the Banach Fixed Point Theorem. See Appendix \ref{proof: convergence} for the complete proof. 
% \end{proof}
% \begin{proof}
% \TODO{put in the appendix. Show it's contracting and use banach fixed-point theorem}
% \textcolor{red}{Kenji: I added the proof in the appendix.}
% \end{proof}
\paragraph{Multiscale propagation} 
With a set of multiple scales $M = \{m_1, ..., m_k | m_i \neq m_j \forall i, j\}$, we can have multiple propagation modules and obtain $k$ equilibriums with different scales $\{Z^{*1}, Z^{*2}, ..., Z^{*k}\}$. 
Given those equilibriums, we propose a scale-aggregation mechanism utilizing learnable attentions, as to learn the contributions of different scales for each node automatically through the learning objective. 
For each node $i$, $z_i^{*t}$ denotes $t$-th equilibrium in $Z^{*t}$ and the attention value $\beta^{t}_i$ is defined as follows: 
\begin{equation}
\label{eq:attention}
        \beta_i^{t} = q^T \text{tanh} (W_{a}z_{i}^{*t} + b_{a}), 
\end{equation}
where $q$ is the parameterized attention weight vector, $W_{a}$ and $b_{a}$ are the weight matrix and the bias vector, respectively. 
Given attention values for different scales $\{\beta^1, ... \beta^k\}$, the final weights are normalized by softmax function: 
\begin{equation}
        \alpha_i^t = \text{softmax}(\beta_i^t) = \frac{\exp(\beta_i^t)}{\sum_{j=1}^k \exp(\beta_i^j)}.
\end{equation}
Larger $\alpha^t_i$ indicates that the corresponding scale is more important for node $i$. 
The final embeddings $Z'$ are obtained by aggregating the equilibriums at different scales with corresponding weights: 
\begin{align}
    z'_i &= \sum_{t=1}^k \alpha_i^t z^{*t}, \\
    \hat Y &= f_{o}(Z'), 
\end{align}
where the predictions $\hat Y$ are generated by a problem-specific decoding function $f_{o}$. 

% \begin{theorem}
% \label{theorem:convergence}
% Given the bounded damping factor $\gamma \in (0, 1)$, the proposed iterative map for propagation (i.e., Equation \eqref{eq:multiscale_prop}) is a contraction mapping and the unique fixed-point solution $Z^{*}$ can be obtained by iterating Equation \eqref{eq:multiscale_prop}.  
% \end{theorem}
% \begin{proof}

% \end{proof}
% \TODO{Proof about the convergence} 

\subsection{Expanded range via multiscale propagation}
In multiscale propagation, nodes receive information from further $m$-hop neighbors rather than only immediate neighbors, which enlarges the effective range of message passing. 
It is similar with a larger receptive field in convolutional neural networks. 
We prove that the effective range for receiving distant information is enlarged by using multiscale propagation. 
\begin{theorem}
\label{new sensitivity}
Given two nodes $p$ and $q$ are $h$-hop apart, using propagation with m-hop neighbors (i.e., Equation \eqref{eq:multiscale_prop}), if we perturb node features $X_{:,p}$ of node $p$ by $\Delta X_{:,p} \in \mathbb{R}^{m}$, the L2 norm of the change in node $q$'s equilibrium $\|\Delta Z^{*}_{:, q}\|$ is upper bounded as follows: 
\begin{equation}
\label{eq:gradients}
%   \Delta Z^{*}_{:, q} < (h_{\theta} - \frac{h}{m})\gamma^{\frac{h}{m}}g^{\frac{h}{m}}(F)(\Delta X_{:,p}S^{h}_{p,q}), 
%    \Delta Z^{*}_{:, q} < \frac{\epsilon^{\frac{1}{m}}}{1 - \gamma}g^{\frac{h}{m}}(F)(\Delta X_{:,p}S^{h}_{p,q})
    \|\Delta Z^{*}_{:, q}\| \leq \frac{\gamma^{\frac{h}{m}}}{1 - \gamma} \|g^{\frac{h}{m}}(F)\Delta X_{:,p}S^{h}_{p,q}\|.
\end{equation}
%assuming $\gamma^i \leq \epsilon$ for all $i \geq h$.
\end{theorem}
The complete proof is provided in Appendix \ref{proof of expanded range}.
% \begin{proof}
% \TODO{}
% \end{proof}

Similar to Corollary \ref{corollary: limited effective range}, we analyse the effective range of multiscale propagation by considering when the change in the equilibrium of node $q$ becomes smaller than a certain numerical error. 
\begin{corollary}
\label{corollary:expanded range}
Using propagation with $m$-hop neighbors (i.e., Equation \eqref{eq:multiscale_prop}, given any small error constant $\theta$, the $\theta$-effective range 
$h < \frac{m\ln (\theta(1-\gamma))}{\ln \gamma}$. Hence, the perturbation on node features of node $p$ can affect the equilibrium of nodes which are located up to $\frac{m\ln (\theta(1-\gamma))}{\ln \gamma}$-hop away from $p$.
\end{corollary}
This is the corollary of Theorem \ref{new sensitivity}. The proof is given in Appendix \ref{proof of expanded range}. Under the same condition, the effective range is expanded by using multiscale propagation to consider $m$-hop neighbors in a propagation step. 
% The change on node $q$'s equilibrium $\Delta Z^{*}_{:,q}$ vanishes to zero when the distance between node q and p reachs $mh_{\theta}$.

\subsection{Training \ourmodel}
% \TODO{To briefly mention how to train the model? Just like sec 4.3 in IGNN and sec 5 in CGS. Probably the reviewers are not familiar with this. Then will have the question about how to train it in practice? }
To train \ourmodel, we can simply iterate Equation \eqref{eq:multiscale_prop} until it converges to the equilibrium $Z^{*}$ for the forward pass. 
We do not use closed-form solutions as in \citet{EIGNN} since it would slow the training with large graphs and a large number of node features. 
For backward pass, given a loss $\ell$, we can use implicit differentiation to compute the gradients of trainable parameters by directly differentiating through the equilibrium $Z^{*}$ by: 
\begin{equation}
\label{eq: gradients}
    \frac{\partial \ell}{\partial (\cdot)} = \frac{\partial \ell}{\partial Z^{*}} \left(I - J_{\varphi}(Z^{*}) \right)^{-1} \frac{\partial \varphi(Z^{*},X,\mathcal{G})}{\partial (\cdot)},
\end{equation}
where $Z^{*} = \varphi(Z^{*},X,\mathcal{G}) = \gamma g(F)Z^{*}S^m + f(X, \mathcal{G})$ and $J_{\varphi}(Z^{*}) = \frac{\partial \varphi(Z^{*},X,\mathcal{G})}{\partial Z^{*}}$. We provide the complete derivation in Appendix \ref{derivation of gradients}. 
One of advantages of directly differentiating through $Z^{*}$ is that the memory consumption is only one layer regardless the number of iterative steps in forward pass. 
In contrast, differentiating over iterative steps requires large memory to store intermediate variables. 

Note that $\left(I - J_{\varphi}(Z^{*})\right)^{-1}$ in Equation \ref{eq: gradients} is expensive to compute due to the computation of the Jacobian $J_{\varphi}(Z^{*})$ and the inverse. 
% Note that the Jacobian $J_{\varphi}(Z^{*})$ and the matrix inverse are expensive to compute. 
However, we can solve a linear equation with a Vector-Jacobian product (VJP) to achieve cheaper computation of $\frac{\partial \ell}{\partial Z^{*}}\left(I - J_{\varphi}(Z^{*})\right)^{-1}$: 
\begin{equation}
    u^T = u^TJ_{\varphi}(Z^{*}) + \frac{\partial \ell}{\partial Z^{*}}.
\end{equation}
Note that, the VJP $u^T J_{\varphi}(Z^{*})$ can be efficiently computed by automatic differentiation packages (e.g., PyTorch \citep{NEURIPS2019_9015}) without forming the Jacobian. 
Subsequently, the gradients $\frac{\partial \ell}{\partial (\cdot)}$ can be obtained. 
% After that, the gradients of any parameters $\frac{\partial \ell}{\partial (\cdot)}$ can be easily computed by existing automatic differentiation packages (e.g., PyTorch).

% which can be represented with $Z*$ as the intermediate variable: 
% \begin{equation}
%     \frac{\partial \ell}{\partial (\cdot)} = \frac{\partial \ell}{\partial Z^{*}}\frac{\partial Z^{*}}{\partial (\cdot)} 
% \end{equation}

% we resort implicit differentiation to obtain gradients by differentiate at the equilibrium  
\subsection{Discussion and comparison with previous implicit GNNs}
% \TODO{Discussion about the difference between our model and CGS (especially the multi-head part).}
% Previous implicit GNNs mainly fall into two categories: 1) use iterative solvers to obtain fixed-point solutions as in IGNN \citep{IGNN} and CGS \citep{CGS}; 2) derive a closed-form solution 
In general, compared to previous implicit GNNs (i.e., IGNN \citep{IGNN}, EIGNN \citep{EIGNN}, and CGS \citep{CGS}), our model \ourmodel brings multiscale modeling to implicit GNNs, which allows the model to capture graph information of different granularities.
The multiscale idea has shown its effectiveness on explicit GNN models, such as N-GCN \citep{NGCN}, JKNet \citep{JKNet}, and MixHop \citep{mixhop}. 
However, to our knowledge, there is no previous implicit GNNs able to capture graph information at various scales.
Specifically, \ourmodel essentially has multiple implicit layers with different focuses on various scales coexisting side by side. 
% Specifically, \ourmodel essentially has different implicit layers focusing on various scales coexisting side by side. 
% whereas CGS parallel multiple implicit layers with the same scale (i.e., only consider immediate neighbors) as the multi-head extension. 

\paragraph{Computational complexity}
\ourmodel has similar time complexity with IGNN and CGS as they all use iterative methods to iterate Equation \eqref{eq:multiscale_prop} until convergence. The asymptotic time complexity is $O(K(h^2n+hn^2))$ where $h$ is the number of hidden units after the input transformation $f(X, \mathcal{G})$ and $K$ is the number of iterations in an iterative method. 
In contrast, EIGNN costs $O(n^3)$ to conduct eigendecomposition for the adjacency $S$, which is costly and prohibitive for large graphs. Additionally, for training, it requires $O(h_i^3 + h_i^2n)$ to get the closed-form solution and $O(h_i^3)$ to conduct eigendecomposition for the weights, where $h_i$ is the number of input features.
Comparing \ourmodel and EIGNN, 
% aside from multiscale modeling,
\ourmodel is more efficient on large graphs as it utilizes iterative solvers for fixed-point solutions, whereas EIGNN requires eigendecomposition to get closed-form solutions. 
% \ourmodel has similar time complexity with IGNN and CGS as they all use iterative methods to iterate Equation \eqref{eq:multiscale_prop} until convergence. The asymptotic time complexity is $O(K(m^2n+mn^2))$ where $K$ is the number of iterations in an iterative method. 
% In contrast, EIGNN costs $O(n^3)$ to conduct eigendecomposition for the adjacency $S$, which is very costly and prohibitive for large graphs. Additionally, for training, it requires $O(m^3 + m^2n)$ to get the fixed-point solution and $O(m^3)$ to conduct eigendecomposition for the weight matrix.
% Regarding the efficiency of iterative methods, \citet{bai2022neural} propose a neural solver to accelerate the iterative method used in implicit models on image and text data. 

The above analysis of computational complexity mainly considers the process of propagation. Here, we discuss the time complexity of some additional operations in these implicit GNNs, e.g., the attention mechanism in MGNNI and the projection on the weight matrix in IGNN, which generally cost much less time compared with the main propagation process. The attention mechanism used in MGNNI has the time complexity $O(h'hn+hn)$ (according to Equation \ref{eq:attention}), where $h'$ is the number of hidden units in the attention module. Similarly, IGNN also has some additional operations, e.g., it requires a projection of the weight matrix $W$ in each training iteration to ensure the well-posedness condition $\left\lVert W\right\rVert_{\infty} \leq \kappa / \lambda_{\mathrm{pf}}(A)$. It needs $O(n^2)$ to get and modify the maximum row sum of $W$ (i.e., $\left\lVert W \right\rVert_{\infty}$). 
% And this operation may not be easily optimized by GPU (see IGNN official implementation \footnote{ https://github.com/SwiftieH/IGNN/blob/main/nodeclassification/utils.py\#L}). 
Overall, MGNNI has the same level of time complexity compared with IGNN and CGS.
% For multiscale propagation in MGNNI, it would linearly increase the complexity of propagation from $O(K(h^2n+hn^2))$ to $O(k'K(h^2n+hn^2))$ if we use $k'$ different scales.
% However, $K$ is the number of iterations in an iterative method (usually $K$ is around 100 in IGNN, CGS, and MGNNI), thus 
% And this operation may not be easily optimized by GPU (see IGNN official implementation \footnote{ https://github.com/SwiftieH/IGNN/blob/main/nodeclassification/utils.py}).
% Multiscale Propagation would increase the complexity of $O(K(h^2n+hn^2))$ to $O(k'K(h^2n+hn^2))$ if we use $k'$ different scales. However, remind that $K$ is the number of iterations in an iterative method (usually $K$ is around 100 in IGNN, CGS, and MGNNI). Thus, we believe this increase by $k'$ would not increase the whole time complexity much considering we usually use $k' \leq 3$ in our experiments.

% \end{equation}
\section{Experiments}
In this section, we show that \ourmodel can effectively capture long-range dependencies and mixed graph information at various scales. Therefore, \ourmodel provides better performance on both node classification and graph classification compared with representative baselines. 
Specifically, we conduct the experiments\footnote{The implementation can be found at https://github.com/liu-jc/MGNNI} on 7 datasets for node classification (including 1 synthetic and 6 real-world datasets: Color-counting, Cornell, Texas, Wisconsin, Chameleon, Squirrel, PPI) and 6 datasets (MUTAG, PTC, PROTEINS, NCI1, IMDB-Binary, IMDB-Multi) for graph classification. 
As we follow the same experimental settings on some datasets, we reuse the results of some baselines from literatures. 
% TODO: Describe the model choices
Descriptions of datasets and experimental settings are detailed in Appendix \ref{appendix: details of experiments}.

% \TODO{Some analyses or experiments we may add after the main results: 1) attention value analysis on some datasets (hopefully we can show different nodes assign different values on different scales); 2)}
\subsection{Experiments with synthetic datasets}
% \paragraph{Chains dataset} 
% \TODO{maybe remove the chain dataset as EIGNN is already 100\%}
% We conduct the experiments on the Chains dataset to directly examine the models' abilities for capturing long-range dependencies as in \citet{IGNN}.
% The graph is constructed by several disjointed chains, where each chain has $l$ nodes and the class information is only sparsely encoded in the feature of the staring end node of each chain. 
% The goal is to classify nodes and the nodes on the same chain have the same label as the starting end node. 

% Results... 
\begin{wrapfigure}{r}{0.4\textwidth}
   	\centering
	\includegraphics[width=0.4\textwidth]{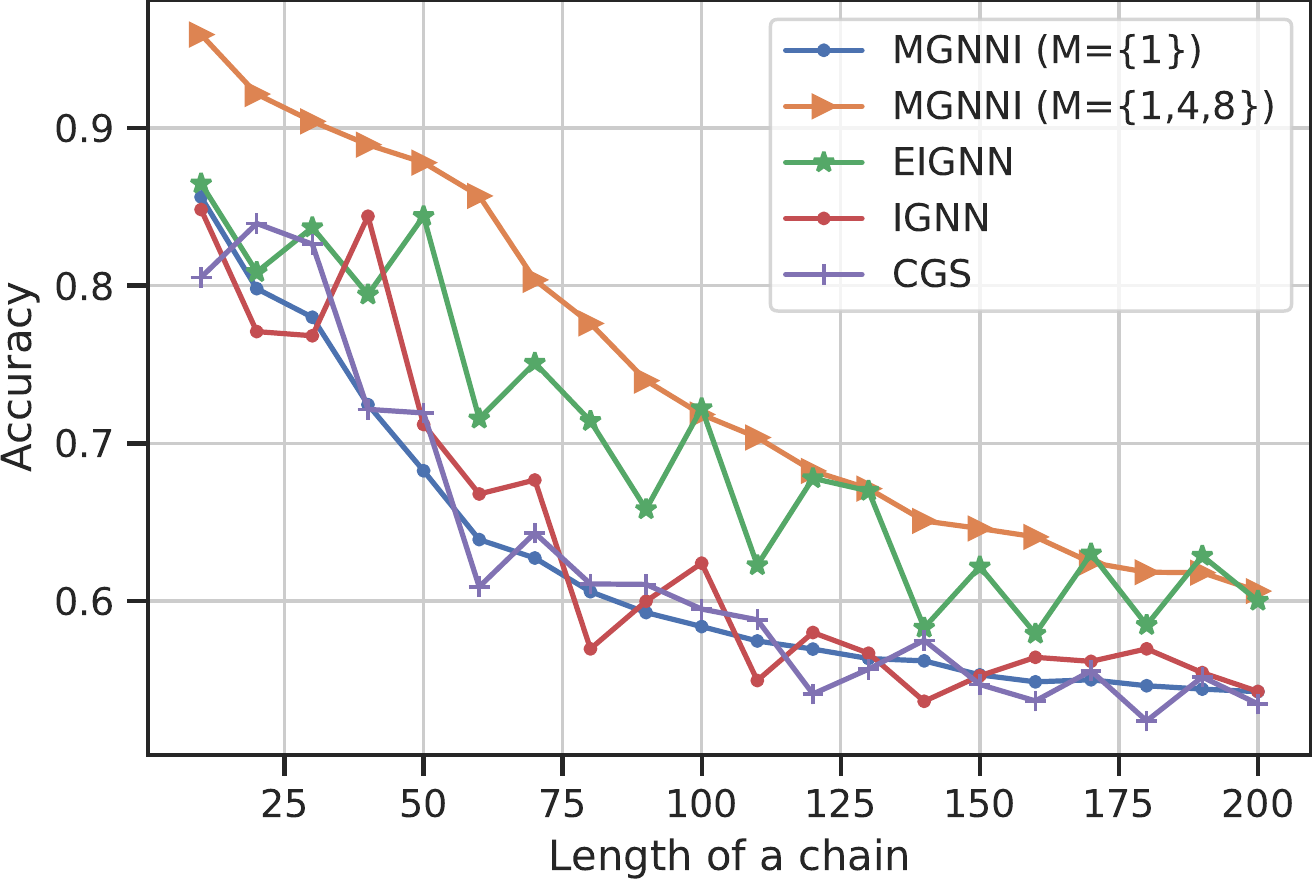}
	\caption{Averaged accuracies on color-counting dataset.}
	\label{fig: color-counting}
% 	\caption{Averaged accuracies with respect to the length of chains.}
    \vspace{-2.0mm}
\end{wrapfigure}
\paragraph{Color-counting dataset}
We construct the synthetic dataset for node classification. The graph contains several chains and some nodes on each chain have different colors.
The color information is encoded in node features. 
The nodes on the same chain share the same label which is the majority color appeared on this chain. 
The model is supposed to predict the majority color on a chain, which requires the model able to capture the long-range dependencies and also count the occurrence of each color. 
% \begin{figure}
% 	\begin{subfigure}[b]{0.5\textwidth}
% 		\centering
% 		\includegraphics[width=1\textwidth]{Figs/binary-chains.pdf}
% 		\caption{Results on plain chain dataset}
% 		\label{fig: binary-chains}
% 	\end{subfigure}
% 	\begin{subfigure}[b]{0.5\textwidth}
% 		\centering
% 		\includegraphics[width=1\textwidth]{Figs/binary-counting-chains.pdf}
% 		\caption{Results on color-counting dataset.}
% 		\label{fig: color-counting}
% 	\end{subfigure}
% 	\caption{Averaged accuracies with respect to the length of chains.}
% 	\label{fig: overall-chains-results}
% \end{figure}
In Figure \ref{fig: color-counting}, we compare \ourmodel with different scales against previous implicit GNN models (i.e., IGNN, EIGNN, and CGS).
% For \ourmodel, different scale combinations are considered, e.g., 
\mbox{\ourmodel (M=\{1,4,8\})} denotes that three scales with m=1, m=4, and m=8 are used.
\ourmodel with higher exponents (i.e., M=\{1,4,8\}) generally performs better than IGNN and CGS, which indicates that multiscale propagation with higher-hop neighbors can effectively expand the effective range for long-range dependencies.
% As IGNN performs the worst, it verifies that previous implicit GNNs indeed have the constrained expressiveness.
% The inferior performance of IGNN and CGS demonstrates that previous implicit GNNs indeed have the constrained expressiveness caused by the .
% Besides, note that \ourmodel (M=\{1\}) can be treated as the EIGNN model but with iterative solvers. 
Note that \ourmodel (M=\{1\}) can be regarded as \ourmodel with only single scale. It is outperformed by \ourmodel (M=\{1,4,8\}), which again shows the effectiveness of multiscale propagation.
\mbox{\ourmodel (M=\{1\})}, IGNN and CGS are all outperformed by EIGNN, suggesting that implicit GNNs with iterative solvers may still suffer from approximation error issues.
% Comparing \ourmodel (m=[1,4,8]) and EIGNN, EIGNN generally achieves worse performance when the chain length < 160, whereas EIGNN outperform \ourmodel (m=[1,4,8]) with the chain length more than 160.
We also provide the comparison between \ourmodel and representative explicit GNNs in Appendix \ref{appendix:synthetic_exp} (See Figure \ref{fig:counting_explicitGNNs}). 
It further demonstrates the superiority of \ourmodel over explicit GNNs in terms of the ability of capturing long-range dependencies. 
% \TODO{Mention that we have more results for comparing MGNNI with explicit GNNs. }

Besides the experiments on Color-counting dataset, following \citet{EIGNN}, we also conduct synthetic experiments on Chains dataset. MGNNI and EIGNN both maintain 100\% test accuracy, demonstrating the advantages of capturing long-range dependencies. The results and detailed analyses can be found in Appendix \ref{appendix:synthetic_exp} and Figure \ref{fig:chain_bincls}.
\subsection{Experiments with real-world datasets} 
\paragraph{Node classification} 
Apart from the synthetic experiments, following \citet{Pei2020Geom-GCN}, we also conduct experiments on 5 heterophilic datasets (Cornell, Texas, Wisconsin, Chameleon, and Squirrel) for node classification.
% Following \citet{Pei2020Geom-GCN}, we use 5 heterophilic graph datasets (Cornell, Texas, Wisconsin, Chameleon, Squirrel) to evaluate the model ability of capturing long-range dependencies. 
On heterophilic graphs, the nodes with different class labels tend to be connected, which requires models to aggregation information from distant nodes. 
These graph datasets are web-page graphs of the corresponding universities or the corresponding Wikipedia pages. 
We use the standard train/test/val splits as in \citet{Pei2020Geom-GCN}.
See the detailed setting in Appendix \ref{appendix:node_cls_real_world}.

\begin{table}[]
	\caption{Results on heterophilic graph datasets: mean accuracy (\%) $\pm$ stdev over different data splits.
% 	``*" denotes that the results are obtained from \citep{Pei2020Geom-GCN}.
	}
	\label{tab: heterophilic results}
% 	\vspace{1mm}
	\centering
	\begin{tabular}{@{}lccccc@{}}
		\toprule
		& \textbf{Cornell} & \textbf{Texas} & \textbf{Wisconsin} & \textbf{Chameleon} & \textbf{Squirrel} \\
		\textbf{\# Nodes}   & 183     & 183   & 251       & 2,277     & 5,201    \\
		\textbf{\# Edges}   & 280     & 295   & 466       & 31,421    & 198,493  \\
		\textbf{\# Classes} & 5       & 5     & 5         & 5         & 5        \\ \midrule
		Geom-GCN \citep{Pei2020Geom-GCN} & 60.81   & 67.57 & 64.12     & 60.90     & 38.14    \\
		SGC \citep{SGC}     & 58.91 $\pm$ 3.15   & 58.92 $\pm$ 4.32 & 59.41 $\pm$ 6.39     & 40.63 $\pm$ 2.35     & 28.4 $\pm$ 1.43      \\
		GCN \citep{semi_GCN}     & 59.19 $\pm$ 3.51   & 64.05 $\pm$ 5.28 & 61.17 $\pm$ 4.71     & 42.34 $\pm$ 2.77     & 29.0 $\pm$ 1.10     \\
		GAT \citep{GAT}     & 59.46 $\pm$ 6.94   & 61.62 $\pm$ 5.77 & 60.78 $\pm$ 8.27     & 46.03 $\pm$ 2.51     & 30.51 $\pm$ 1.28    \\
		APPNP \citep{APPNP}   & 63.78 $\pm$ 5.43   & 64.32 $\pm$ 7.03 & 61.57 $\pm$ 3.31     & 43.85 $\pm$ 2.43     & 30.67 $\pm$ 1.06    \\
		JKNet \citep{JKNet}   & 58.18 $\pm$ 3.87   & 63.78 $\pm$ 6.30  & 60.98 $\pm$ 2.97     & 44.45 $\pm$ 3.17     & 30.83 $\pm$ 1.65    \\
		GCNII \citep{GCNII}   & 76.75 $\pm$ 5.95   & 73.51 $\pm$ 9.95 & 78.82 $\pm$ 5.74     & 48.59 $\pm$ 1.88     & 32.20 $\pm$ 1.06    \\
		H2GCN \citep{H2GCN}   & 82.22 $\pm$ 5.67   & 84.76 $\pm$ 5.57     & 85.88 $\pm$ 4.58     & 60.30 $\pm$ 2.31     & 40.75 $\pm$ 1.44    \\ 
		\midrule
		IGNN \citep{IGNN}    & 61.35 $\pm$ 4.84   & 58.37 $\pm$ 5.82 & 53.53 $\pm$ 6.49     & 41.38 $\pm$ 2.53     & 24.99 $\pm$ 2.11    \\
    	EIGNN \citep{EIGNN} & 85.13 $\pm$ 5.57   &
    	84.60 $\pm$ 5.41 & \textbf{86.86 $\pm$ 5.54 }    & 62.92 $\pm$ 1.59     & 46.37 $\pm$ 1.39 \\
    	CGS \citep{CGS} & 68.11 $\pm$  9.41 & 62.97 $\pm$  9.23 & 63.53 $\pm$ 9.81 & 40.57 $\pm$ 1.61& 31.78 $\pm$ 0.89 \\ 
    	\midrule
		\ourmodel & \textbf{85.95 $\pm$ 6.10} & \textbf{84.86 $\pm$ 5.91} & 86.67 $\pm$ 4.31 & \textbf{63.93 $\pm$ 2.21} & \textbf{54.50 $\pm$ 2.10} \\
		\bottomrule
	\end{tabular}
% 	\vspace{-2mm}
\end{table}

The results are shown in Table \ref{tab: heterophilic results}. 
% As we follow the exact setting as in \citet{EIGNN}, we reuse the results of baselines except CGS. 
MGNNI generally achieves the best performance on most datasets. 
Comparing MGNNI and EIGNN, MGNNI provides better results, especially on Chameleon and Squirrel, which indicates that MGNNI has the superior ability of capturing long-range dependencies and multiscale information. 
Among implicit GNN baselines, EIGNN outperforms IGNN and CGS, which might be attributed to the less approximation error in the closed-form solution of EIGNN.
Among explicit GNN baselines, H2GCN and GCNII are usually better than others, suggesting that the aggregation design in H2GCN and deeper models with residual connections are helpful on heterophilic graphs.

\begin{wraptable}{r}{0.35\textwidth}
    \centering
    \caption{Multi-label node classification on PPI: Micro-F1 (\%).}
    \label{tab: PPI results}
    \begin{tabular}{@{}lc@{}}
    \toprule
    Method                 & Micro-F1      \\ \midrule
    GCN \citep{semi_GCN}                   & 59.2          \\
    GraphSAGE \citep{Graphsage}             & 78.6          \\
    SSE \citep{pmlr-v80-dai18a}                   & 83.6          \\
    GAT \citep{GAT}                   & 97.3          \\
    JKNet \citep{JKNet}                 & 97.6 \\ 
    \midrule
    IGNN \citep{IGNN}                   & 97.6          \\ 
    EIGNN \citep{EIGNN}                 & 98.0 \\ \midrule
    \ourmodel & \textbf{98.7} \\
    \bottomrule
    \end{tabular}
\end{wraptable}
In addition, we also evaluate \ourmodel on Protein-Protein Interaction (PPI) dataset which have multiple graphs. On each graph, proteins are presented as nodes and edges are formed if there is an interaction between two proteins. The task is to predict node labels on multi-label multi-graph inductive setting. The same train/val/test split are used as in \citet{Graphsage}. 
Table \ref{tab: PPI results} reports the micro-F1 scores of \ourmodel against other baseline models. 
Compared to IGNN and EIGNN, \ourmodel achieves 1.1\% and 0.7\% absolute improvement respectively by effectively capturing underlying multiscale information and long-range dependencies between proteins. 
Unlike IGNN having 4 implicit layers sequentially stacked, on PPI dataset, \ourmodel resorts to parallel equilibrium layers with different scales, which makes \ourmodel more efficient than IGNN. 
We report the efficiency comparison between \ourmodel and other implicit GNNs in Appendix \ref{app:efficiency_comparsion}.
% As shown in Table \ref{tab: training time PPI}, \ourmodel only spends slightly more time than EIGNN for training an epoch, meanwhile, \ourmodel does not require eigendecomposition on adjacency matrix as the pre-process which can be very costly if the graph size is larger. 
% Besides the performance comparison, we also compare \ourmodel with other implicit GNNs in terms of efficiency. 

% \begin{wraptable}{r}{0.4\textwidth}
%     \label{tab:PPI_efficiency}
%     \centering
%     \caption{Training time per epoch on PPI.}
%     \label{tab: training time PPI}
%     \begin{tabular}{@{}lcc@{}}
%     \toprule
%     Method                 & Train Time & Pre-process     \\ \midrule
%     IGNN                 & 32.7s  & N.A.        \\ 
%     EIGNN               & 2.3s & 45s\\ \midrule
%     \ourmodel & 2.6s & N.A. \\
%     \bottomrule
%     \end{tabular}
% \end{wraptable}

\begin{table}[]
	\caption{Mean accuracy (\%) $\pm$ stdev over 10 folds on real-world datasets for graph classification. 
% 	Results are obtained from \citet{IGNN}, except IGNN and CGS. 
	}
% 	\vspace{1mm}
	\label{tab: graph classification real-world results}
	% how to add a line space between title and table 
	\centering
% 	\resizebox{0.75\textwidth}{!}{
	\begin{tabular}{@{}lcccccc@{}}
		\toprule
		& \textbf{MUTAG} & \textbf{PTC} & \textbf{PROTEINS}  & \textbf{NCI1} & \textbf{IMDB-B} & \textbf{IMDB-M} \\
		\textbf{\# Graphs}   & 188     & 344   & 1113 & 4110 & 1000 & 1500        \\
        \textbf{Avg \# Nodes } & 17.9 & 25.5 & 39.1 & 29.8 & 19.8 & 13.0 \\
		\textbf{\# Classes} & 2  & 2 & 2 & 2 & 2 & 3            \\
		\midrule
		GCN \citep{semi_GCN} & 85.6 $\pm$ 5.8 & 64.2 $\pm$ 4.3 & 76.0 $\pm$ 3.2 & 80.2 $\pm$ 2.0 & - & -\\ 
		GIN \citep{GIN} &89.0 $\pm$ 6.0 & 63.7 $\pm$ 8.2 & 75.9 $\pm$ 3.8 & \textbf{82.7 $\pm$ 1.6} & 75.1 $\pm$ 5.1 & 52.3 $\pm$ 2.8 \\
		DGCNN \citep{zhang2018end} & 85.8 & 58.6 & 75.5 & 74.4 & 70.0 & 47.8 \\ 
		FDGNN \citep{gallicchio2020fast} & 88.5 $\pm$ 3.8 & 63.4 $\pm$ 5.4 & 76.8 $\pm$2.9 & 77.8 $\pm$ 1.6 & 72.4 $\pm$ 3.6 & 50.0 $\pm$ 1.3\\ 
		\midrule
		IGNN \citep{IGNN} & 89.3 $\pm$ 6.7 & 70.1 $\pm$ 5.6 & 77.7 $\pm$ 3.4 & 80.5 $\pm$ 1.9 & - & - \\
		EIGNN \citep{EIGNN} & 88.9 $\pm$ 1.1& 69.8 $\pm$ 5.3 & 75.9 $\pm$ 6.4 & 77.5 $\pm$ 2.2 & 72.3 $\pm$ 4.3 & 52.1 $\pm$ 2.9\\
		CGS \citep{CGS} & 89.4 $\pm$ 5.6 & 64.7 $\pm$ 6.4 & 76.3 $\pm$ 6.3 & 77.2 $\pm$ 2.0 & 73.1 $\pm$ 3.3 & 51.1 $\pm$ 2.2 \\
		\midrule
		\ourmodel & \textbf{91.9 $\pm$ 5.5} & \textbf{72.1 $\pm$ 2.8} &  \textbf{79.2 $\pm$ 2.9} & 78.9 $\pm$ 2.1 & \textbf{75.8 $\pm$ 3.4} & \textbf{53.5 $\pm$ 2.8}\\
		\bottomrule
	\end{tabular}
% 	}
\vspace{-0.2cm}
\end{table}
\paragraph{Graph classification}
Besides node classification, we conduct experiments on graph classification tasks, using four bioinformatics datasets (MUTAG, PTC, PROTEINS, NCI) \citep{yanardag2015deep} and two social-network datasets (IMDB-Binary and IMDB-Multi). 
% See Appendix \ref{appendix:graph_cls} for detailed descriptions and settings.
10-fold Cross-validation is conducted as \citep{GIN} and the averaged accuracies with standard deviations are reported in Table \ref{tab: graph classification real-world results}. The results of baselines are borrowed from \citet{IGNN} and \citet{CGS}.
In general, compared with other implicit and explicit baseline models, \ourmodel achieves the best performance on 3 out of 4 bioinformatics datasets and both two social-network datasets.
This shows that the ability to capture long-range dependencies and multiscale information is helpful for graph classification and can be generalized to test graphs in the inductive setting. 
% explain implicit GNNs generally worse than GIN in NCI1.
\subsection{Ablation Study}
\label{sec: ablation study}
% 1) compare single scale at different datasets (1 vs 2 vs 3). Like immediate neighbors are most important on PPI, while 2- or 3-hop neighbors are more important on Texas. 
% 2) compare single scale with multiscale. incorporating different scales improve the model capacity. 
\begin{wraptable}{r}{0.4\textwidth}
    \centering
    \vspace{-0.3cm}
    \caption{Performance of different scales.}
     \label{tab:ablation study}
    \begin{tabular}{@{}cccc@{}}
        \toprule
         Scales & PPI & Chameleon & Texas \\
         \midrule
         \{1\} & 98.35 & 61.46 & 81.35\\
         \{2\} & 94.62 & 58.24 & 82.97 \\
         \{3\} & 88.56 & 56.07 & 83.78 \\
         \midrule
         \{1,2\} & 98.67 & 63.93 & 83.24\\
         \{1,2,3\} & 98.74 & 63.75 & 84.86 \\ 
        \bottomrule
    \end{tabular}
\end{wraptable}
To further investigate the effectiveness of multiscale modeling and the contributions of different scales, we conduct the ablation study using different \ourmodel variants with various scale combinations. 
As shown in Table \ref{tab:ablation study}, compared to variants with a single scale (i.e., \{1\}, \{2\} and \{3\}), multiscale variants generally achieve better performance on all three datasets.  
This verifies that multiscale modeling is effective and plays an important role in capturing graph information at various scales. 
Comparing variants with a single scale, the variant utilizing 1-hop propagation performs best on PPI and Chameleon, whereas the variant with only 3-hop propagation obtains the best performance on Texas.
This demonstrates that information at a certain scale can be more important than others on graphs with different properties. Merely considering 1-hop propagation as in previous implicit GNNs might lead to sub-optimal performance. 

\begin{wrapfigure}{r}{0.47\textwidth}
    \centering
    \vspace{-0.2cm}
    \includegraphics[width=0.47\textwidth]{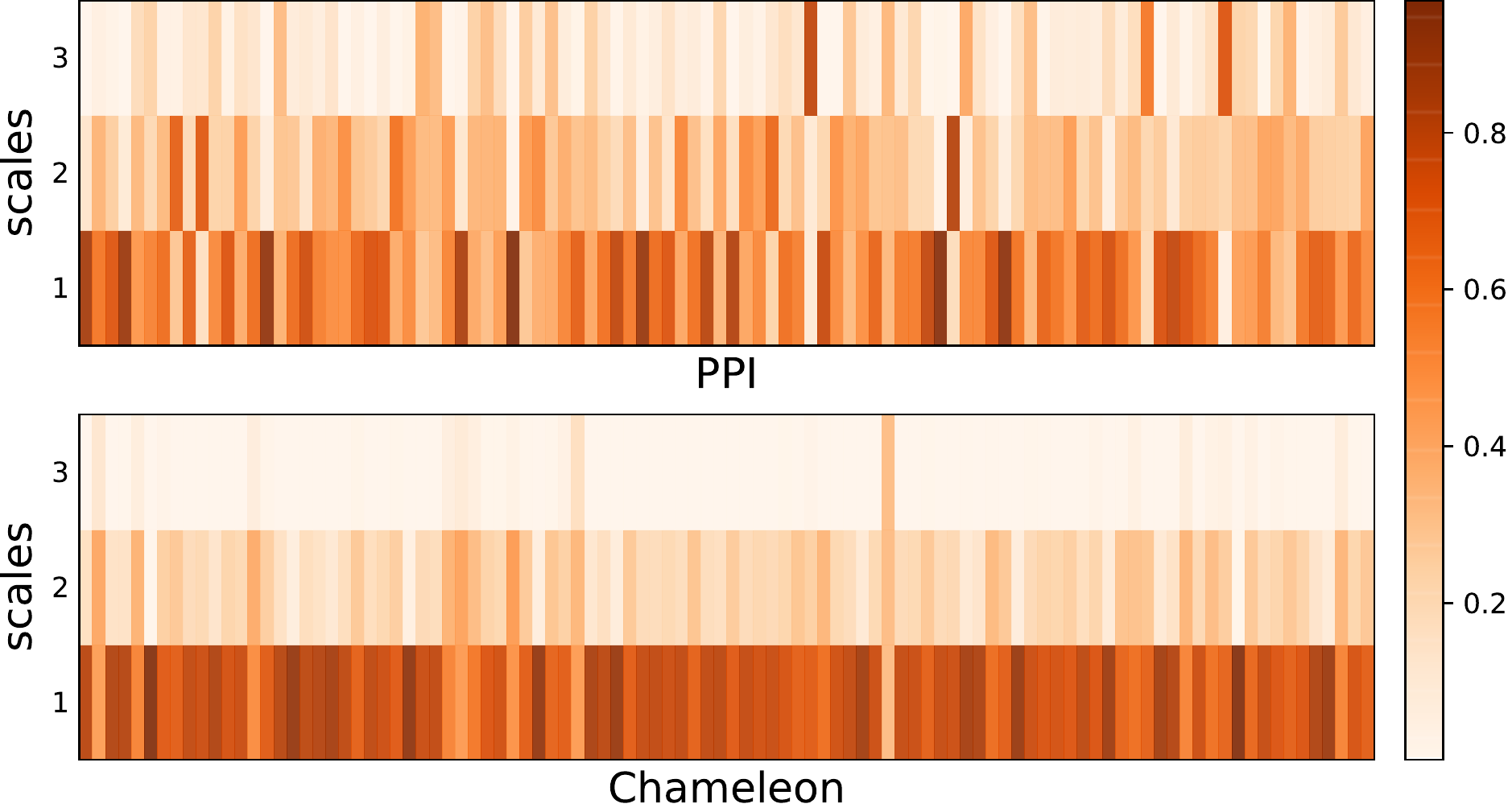}
 	\caption{Attention weights of nodes at different scales on PPI and Chameleon.}
 	\label{fig: att_vals}
    
\end{wrapfigure}
% \TODO{a small figure for attention values to show that the model can learn different scale preferences for  different nodes}
Besides the overall performance of different variants, in Figure \ref{fig: att_vals}, we also demonstrate the attention values of 100 randomly sampled nodes using \ourmodel with scales $\{1,2,3\}$ on PPI and Chameleon. 
Most of the nodes on chameleon tend to have high attention weights on the first scale and rarely have attentions on the third scale,
whereas some nodes on PPI have relatively higher attention values on the third scale than on the first scale.
Moreover, compared to Chameleon, there are more nodes on PPI prefer the second and the third scale. 
These phenomena verify the effectiveness of the attention module in \ourmodel, and show that different nodes might prefer information at different scales according to the graph property. 
Additionally, we also conduct the ablation study on the effect of the attention mechanism in MGNNI (see Appendix \ref{appendix: attention mechanism effect}).

% To quantitatively investigate the effect of the attention mechanism in MGNNI, we conduct additional experiments by removing the attention mechanism and instead use average pooling for fusing information from multiple scales. 
% The experimental results are provided in Table \ref{tab:removing_attention}.

% \begin{wraptable}{r}{0.45\textwidth}
%     \centering
%     \caption{Performance of different scales.}
%      \label{tab:removing_attention}
%     \begin{tabular}{@{}cccc@{}}
%         \toprule
%          Scales & PPI & Chameleon & Texas \\
%          \midrule
%          (wo/ att) \{1,2\} & 98.35 & 61.46 & 81.35\\
%          (w/ att) \{1,2\} & 94.62 & 58.24 & 82.97 \\
%          \midrule
%          (wo/ att) \{1,2\} & 98.67 & 63.93 & 83.24\\
%          (w/ att) \{1,2\} & 98.74 & 63.75 & 84.86 \\ 
%         \bottomrule
%     \end{tabular}
% \end{wraptable}
\section{Conclusion}
In this paper, we provide theoretical analyses on the constrained expressiveness of previous implicit GNNs due to the limited effective range for capturing distant information.
We propose \ourmodel which has an expanded effective range and the ability to learn mixing graph information at various scales.
With synthetic experiments, we show that \ourmodel with a large effective range has a better capacity to capture long-range dependencies.
On various real-world datasets for both node classification and graph classification, \ourmodel demonstrates superior performances compared with representative baselines, showing the effectiveness of multiscale modeling.
Furthermore, the ablation study also shows that \ourmodel allows different nodes to have different scale preferences, which plays an important role in adaptively capturing graph information at various scales. 

% \TODO{limitations.}
% As in IGNN and CGS, share the same approximation error in the iterative solvers? 
% model flexibility? 
Although \ourmodel has the advantages of capturing multiscale graph information, \ourmodel also has the limitation on potential approximation errors introduced in the iterative method as IGNN and CGS. 
EIGNN gets rid of these errors by using a closed-form solution with eigendecomposition which prohibits its usage on large graphs. 
How to mitigate approximation errors while ensuring the scalability on large graphs could be an interesting topic for future work.
\ack 
We would like to gratefully thank the insightful feedback and suggestions from the anonymous reviewers. We also appreciate their engagement during the discussion period. 
This work was supported by Proximate Beta (Grant No. A-8000177-00-00).
The views and conclusions contained in this paper are those of the authors and should not be interpreted as representing any funding agencies. 
% \clearpage
\bibliographystyle{plainnat}
\bibliography{reference}
\clearpage
\section*{Checklist}

\begin{enumerate}

\item For all authors...
\begin{enumerate}
  \item Do the main claims made in the abstract and introduction accurately reflect the paper's contributions and scope?
    \answerYes{}
  \item Did you describe the limitations of your work?
    \answerYes{We discuss the limitations in our conclusion.}
  \item Did you discuss any potential negative societal impacts of your work?
    \answerNA{}
  \item Have you read the ethics review guidelines and ensured that your paper conforms to them?
    \answerYes{}
\end{enumerate}

\item If you are including theoretical results...
\begin{enumerate}
  \item Did you state the full set of assumptions of all theoretical results?
    \answerYes{}
	\item Did you include complete proofs of all theoretical results?
    \answerYes{They are included in Appendix \ref{appendix:proofs}}
\end{enumerate}

\item If you ran experiments...
\begin{enumerate}
  \item Did you include the code, data, and instructions needed to reproduce the main experimental results (either in the supplemental material or as a URL)?
    \answerYes{In Appendix \ref{appendix: details of experiments}}
  \item Did you specify all the training details (e.g., data splits, hyperparameters, how they were chosen)?
    \answerYes{In Appendix \ref{appendix: details of experiments}.}
	\item Did you report error bars (e.g., with respect to the random seed after running experiments multiple times)?
    \answerYes{We report standard deviations.}
	\item Did you include the total amount of compute and the type of resources used (e.g., type of GPUs, internal cluster, or cloud provider)?
    \answerYes{See Appendix \ref{appendix: details of experiments}}
\end{enumerate}

\item If you are using existing assets (e.g., code, data, models) or curating/releasing new assets...
\begin{enumerate}
  \item If your work uses existing assets, did you cite the creators?
    \answerYes{}
  \item Did you mention the license of the assets?
    \answerNA{We only use  public datasets and code.}
  \item Did you include any new assets either in the supplemental material or as a URL?
    \answerYes{We include our code in the supplemental material.}
  \item Did you discuss whether and how consent was obtained from people whose data you're using/curating?
    \answerNA{}
  \item Did you discuss whether the data you are using/curating contains personally identifiable information or offensive content?
    \answerNA{}
\end{enumerate}

\item If you used crowdsourcing or conducted research with human subjects...
\begin{enumerate}
  \item Did you include the full text of instructions given to participants and screenshots, if applicable?
    \answerNA{}
  \item Did you describe any potential participant risks, with links to Institutional Review Board (IRB) approvals, if applicable?
    \answerNA{}
  \item Did you include the estimated hourly wage paid to participants and the total amount spent on participant compensation?
    \answerNA{}
\end{enumerate}

\end{enumerate}
\clearpage
\appendix
\appendixpage
\section{Proofs}
\label{appendix:proofs}
\subsection{Proof of Theorem \ref{decaying sensitivity}}
\label{proof of decaying sensitivity}
\begin{proof}
We denote the perturbed node features as $X'$ and change Equation \eqref{eq:Implicit_propagation} to the following equivalent form: 
\begin{equation*}
    Z^{(k)}=\gamma^{k} g^{k}(F) Z^{(0)} S^{k}+\sum_{i=0}^{k-1} \gamma^{i} g^{i}(F) X' S^{i}.
\end{equation*}
Let $k \rightarrow \infty$, the first term becomes zero as $\gamma < 1$. We further decompose the above equation to 
$
    Z^{(k)}=\sum_{i=0}^{h-1} \gamma^{i} g^{i}(F) X S^{i}+\sum_{i=h}^{k-1} \gamma^{i} g^{i}(F) X S^{i}.
$
Since node $p$ and $q$ are h-hop apart, $S^i_{p,q} = 0$ when $i < h$. Then, we have $Z^{(k)}=\sum_{i=h}^{k-1} \gamma^{i} g^{i}(F) X' S^{i}$. 

Let the perturbed features $X_{:, p}^{\prime}=X_{:, p}+\Delta X_{:, p}$, we have $\left(X^{\prime} S^{i}\right)_{:, q}=\left(X S^{i}\right)_{:, q}+\Delta X_{:, p} S_{p, q}^{i}$.
Then we have the following: 
\begin{align}
    \Delta Z^{(k)}_{:, q} & =\sum_{i=h}^{k-1} \gamma^{i} g^{i}(F)\left(\Delta X_{:, p}S_{p, q}^{i}\right)
\end{align}
Apply the L2 norm on the change $\Delta Z^{(k)}_{:,q}$,
\begin{align}
    \| \Delta Z^{(k)}_{:, q}\| & =\sum_{i=h}^{k-1}  \gamma^{i} \|g^{i}(F)\Delta X_{:, p}S_{p, q}^{i} \| \\
    & \leq (\sum_{i=h}^{k-1} \gamma^{i}) \|g^h(F) \Delta X_{:,p} S^h_{p,q} \|\\ 
    & \leq \frac{\gamma^h - \gamma^k}{1 - \gamma} \|g^{h}(F) \Delta X_{:,p} S^h_{p,q}\|
\end{align}
The last inequality is derived by the sum of a geometric series. 

As $k \rightarrow \infty$, $Z^{*} = \lim_{k \rightarrow \infty} Z^{(k)}$ and $\lim_{k \rightarrow \infty} \gamma^k = 0$ . Then we have the following upper bound: 
\begin{equation}
    \|\Delta Z^{*}_{:, q} \| \leq \frac{\gamma^h}{1 - \gamma} \|g^{h}(F) \Delta X_{:,p} S^h_{p,q}\|
\end{equation}
%Since $\gamma^i \leq \epsilon$ for all $i \geq h$, 
%\begin{equation}
%        \Delta Z^{*}_{:, q} \leq \frac{\epsilon}{1 - \gamma} g^{h}(F) \Delta X_{:,p} S^h_{p,q}
%\end{equation}
\end{proof}

\subsection{Proof of Corollary \ref{corollary: limited effective range}}
\label{proof of limited effective range}
\begin{proof}
% When $\Delta Z^{*}_{:,q} \leq \theta$, the change on node $p$ cannot effect the $Z^{*}_{:,q}$. 
% Combining with Equation \eqref{eq: sensitivity} in Theorem \ref{decaying sensitivity}, we can have $\frac{\epsilon}{1-\gamma} \leq \theta$. 
% As $\gamma^h \leq \epsilon$, we have $h \ln \gamma \leq \ln \epsilon$. 
% With $\frac{\epsilon}{1-\gamma} \leq \theta$ and $h \ln \gamma \leq \ln \epsilon$, the following holds:
% % the following holds:
% \begin{equation}
%     h \geq \frac{\ln (\theta(1-\gamma))}{\ln \gamma}
% \end{equation}
% Therefore, when the distance between node $p$ and $q$ goes beyond $\frac{\ln (\theta(1-\gamma))}{\ln \gamma}$, the change on node $p$'s features cannot affect the  % equilibrium of node $q$.

% When $\Delta Z^{*}_{:,q} \leq \theta$, the change on node $p$ cannot affect the $Z^{*}_{:,q}$. 
% Combining with Equation \eqref{eq: sensitivity} in Theorem \ref{decaying sensitivity}, we have $\frac{\gamma^h}{1-\gamma} \leq \theta$. Since $(1-\gamma)>0$, this is equivalent to $\gamma^h \leq (1-\gamma) \theta$ and thus $h \ln \gamma \leq \ln ((1-\gamma) \theta)$. Since $\ln \gamma <0$, we have that 
% \begin{equation}
%  h \geq \frac{\ln (\theta(1-\gamma))}{\ln \gamma}.
% \end{equation}
% Therefore, when the distance between node $p$ and $q$ goes beyond $\frac{\ln (\theta(1-\gamma))}{\ln \gamma}$, the change on node $p$'s features cannot affect the  equilibrium of node $q$.
Given the $\theta$-effective range $h$, combining Definition \ref{def: effective range} and Equation \eqref{eq: sensitivity} in Theorem \ref{decaying sensitivity}, we have $\theta < \Delta Z^{*}_{:,q} \leq \frac{\gamma^h}{1-\gamma} \|g^{h}(F)\Delta X_{:,p}S^{h}_{p,q}\|$. 
Since $(1-\gamma)>0$, this is equivalent to $\gamma^h > (1-\gamma) \theta$ and thus $h \ln \gamma > \ln ((1-\gamma) \theta)$. Since $\ln \gamma <0$, we have that 
\begin{equation}
 h < \frac{\ln (\theta(1-\gamma))}{\ln \gamma}.
\end{equation}
Therefore, if node features
$X_{:,p}$ of node $p$ are perturbed, the perturbation can only affect the equilibrium of nodes which are up
to $\frac{\ln (\theta(1-\gamma))}{\ln \gamma}$-hop away from $p$.
% 152 ln γ -hop away from pwhen the distance between node $p$ and $q$ goes beyond $\frac{\ln (\theta(1-\gamma))}{\ln \gamma}$, the change on node $p$'s features cannot affect the  equilibrium of node $q$.

\end{proof}
\subsection{Proof of Theorem \ref{theorem:convergence}}
\label{proof: convergence}
\begin{proof}
For any matrix $M \in \RR^{m_1 \times m_2}$,  we define the vectorization of the matrix by $\vect[M]\in \RR^{m_{1}m_2}$, and the Frobenius norm of the matrix by $\|M\|_F$. We define the map $ \varphi$ by $\varphi(Z)=\gamma g(F)ZS^{m} + f(X, \mathcal{G})$. Recall that $Z\in \mathbb{R}^{h\times n}$. We want to show that the map $\varphi$ is contraction. Using the property of the vectorization and the Kronecker  product,
\begin{align*}
\vect[\varphi(Z)]=\gamma \vect[g(F)ZS^{m} ]+\vect[ f(X, \mathcal{G})]=\gamma [(S^{m})\T \otimes g(F)]\vect [Z]+\vect[ f(X, \mathcal{G})].
\end{align*}

Therefore, for any $Z,Z'\in \mathbb{R}^{h\times n}$,
\begin{align*}
\|\varphi(Z) - \varphi(Z')\|_F &= \|\vect[\varphi(Z)] - \vect[\varphi(Z')]\|_2
\\ &=\|\gamma [(S^{m})\T \otimes g(F)](\vect[Z]-\vect[Z'])\|_2
\\ & \le\gamma \|[(S^{m})\T \otimes g(F)]\|_{2}\|\vect[Z]-\vect[Z']\|_2 
\\ & =\gamma \|[(S^{m})\T\|_2 \|g(F)]\|_{2}\|\vect[Z]-\vect[Z']\|_2 
\\ & \le \gamma \|Z-Z'\|_F.
\end{align*}
Since $\gamma \in [0,1)$, this shows that $\varphi$ is a contraction mapping on the metric space $(\RR^{h \times n}, \hat d)$ where $\hat d(Z, Z')=\|Z-Z'\|_F$. Thus, using the  Banach fixed-point theorem, the desired statement of this theorem follows.
\end{proof}

\subsection{Proof of Theorem \ref{new sensitivity}}
We denote the perturbed node features as $X'$ and change Equation \eqref{eq:multiscale_prop} to the following equivalent form: 
\begin{equation*}
    Z^{(k)}=\gamma^{k} g^{k}(F) Z^{(0)} S^{mk}+\sum_{i=0}^{k-1} \gamma^{i} g^{i}(F) X' S^{im}.
\end{equation*}
Following the similar procedure in the proof of Theorem \ref{decaying sensitivity}, as $k \rightarrow \infty$ and $S_{p,q}^i = 0$ when $i<h$, we have 
$Z^{(k)}=\sum_{i=\lfloor h/m \rfloor}^{k-1} \gamma^{i} g^{i}(F) X' S^{im}$. 

Let the perturbed features $X_{:, p}^{\prime}=X_{:, p}+\Delta X_{:, p}$, we have $\left(X^{\prime} S^{i}\right)_{:, q}=\left(X S^{i}\right)_{:, q}+\Delta X_{:, p} S_{p, q}^{i}$.
Then we have the following: 
\begin{align}
    \Delta Z^{(k)}_{:, q} & =\sum_{i=\lfloor \frac{h}{m} \rfloor}^{k-1} \gamma^{i} g^{i}(F)\left(\Delta X_{:, p}S_{p, q}^{im}\right) \\
    % & \leq (\sum_{i=h}^{k-1} \gamma^{i}) S^h_{p,q} g^h(F) \Delta X_{:,p} \\ 
    % & \leq \frac{\gamma^{\frac{h}{m}} - \gamma^k}{1 - \gamma} g^{\frac{h}{m}}(F) \Delta X_{:,p}S^h_{p,q}
\end{align}
Apply the L2 norm on the change $\Delta Z^{(k)}_{:,q}$, 
\begin{align}
    \|\Delta Z^{(k)}_{:, q}\| & =\sum_{i=\lfloor \frac{h}{m} \rfloor}^{k-1} \gamma^{i} \|g^{i}(F)\Delta X_{:, p}S_{p, q}^{im}\| \\
    % & \leq (\sum_{i=h}^{k-1} \gamma^{i}) S^h_{p,q} g^h(F) \Delta X_{:,p} \\ 
    & \leq \frac{\gamma^{\frac{h}{m}} - \gamma^k}{1 - \gamma} \|g^{\frac{h}{m}}(F) \Delta X_{:,p}S^h_{p,q}\|
\end{align}
% The last inequality is derived by the sum of a geometric series. 

As $k \rightarrow \infty$, $Z^{*} = \lim_{k \rightarrow \infty} Z^{(k)}$ and $\lim_{k \rightarrow \infty} \gamma^k = 0$ . Then we have the following upper bound: 
\begin{equation}
    \|\Delta Z^{*}_{:, q}\| \leq \frac{\gamma^{\frac{h}{m}}}{1 - \gamma} \|g^{{\frac{h}{m}}}(F) \Delta X_{:,p}  S^h_{p,q}\|
\end{equation}
% As $\gamma^i \leq \epsilon$ for all $i \geq h$, 
% \begin{equation}
%     \Delta Z^{*}_{:, q} \leq \frac{\epsilon^{\frac{1}{m}}}{1 - \gamma} g^{h}(F) \Delta X_{:,p} S^h_{p,q}
% \end{equation}
\subsection{Proof of Corollary \ref{corollary:expanded range}}
% Similarly to the proof of Corollary \ref{corollary: limited effective range} above, if $h$ satisfies that $\frac{\gamma^{h/m}}{1-\gamma} \ge \theta$, then the numerical error does not dominate. Since $(1-\gamma)>0$ and $\ln \gamma <0$, this is equivalent to 
% \begin{equation}
%  h \le \frac{m \ln (\theta(1-\gamma))}{\ln \gamma}.
% \end{equation}
% Therefore, when the distance between node $p$ and $q$ is within $\frac{m \ln (\theta(1-\gamma))}{\ln \gamma}$, the change on node $p$'s features can affect the  equilibrium of node $q$.
\label{proof of expanded range}
\begin{proof}
Similarly to the proof of Corollary \ref{corollary: limited effective range} above, if $h$ satisfies that $\frac{\gamma^{h/m}}{1-\gamma} > \theta$, then the numerical error $\theta$ does not dominate. Since $(1-\gamma)>0$ and $\ln \gamma <0$, this is equivalent to 
\begin{equation}
 h < \frac{m \ln (\theta(1-\gamma))}{\ln \gamma}.
\end{equation}
Therefore, the change on node $p$'s features can affect the equilibrium of node $q$ which is up to $\frac{m \ln (\theta(1-\gamma))}{\ln \gamma}$-hop away from $p$.
\end{proof}

\subsection{Derivation of the gradients}
\label{derivation of gradients}
% Recall that we obtain the equilibrium $Z^{*}$ by iterating the equation $Z^{*} = \varphi($ 
Here, we provide the derivation of Equation \eqref{eq:gradients} for obtaining gradients of trainable parameters with implicit differentiation
Using the chain rule, the gradients of trainable parameters can be computed by:
\begin{equation}
\label{eq:gradients_w_z*}
    \frac{\partial \ell}{\partial (\cdot)} = \frac{\partial \ell}{\partial Z^{*}}\frac{\partial Z^{*}}{\partial (\cdot)},
\end{equation}
where $(\cdot)$ denotes any trainable parameters within or before the implicit layer $Z^{*} = \varphi(Z^{*}, X, \mathcal{G})$. 
Note that $\frac{\partial \ell}{\partial (\cdot)}$ is directly handled by automatic differentiation (autodiff) packages, while $\frac{\partial Z^{*}}{\partial (\cdot)}$ cannot be directly obtained with autodiff since $Z^{*}$ and $(\cdot)$ are implicitly related. 

By differentiating the both side of the fix-point equation, we can have: 
\begin{equation}
    \frac{\partial Z^{*}}{\partial (\cdot)} = \frac{\partial \varphi(Z^{*}, X, \mathcal{G})}{\partial Z^{*}}\frac{\partial Z^{*}(\cdot)}{\partial (\cdot)} + \frac{\partial \varphi(Z^{*}, X, \mathcal{G})}{\partial (\cdot)},
\end{equation}
where we use $Z^{*}(\cdot)$ to denote the case where we treat $Z^{*}$ as an implicit function of variables we are differentiating with respect to (e.g., the parameters of $\varphi$) and $Z^{*}$ alone to refer to the value of equilibrium. 

By rearranging the above equation, we can obtain the explicit expression for $\frac{\partial Z^{*}}{\partial (\cdot)}$: 
\begin{equation}
\label{eq:Z*_derivative}
    \frac{\partial Z^{*}}{\partial (\cdot)} = \left(I - J_{\varphi}(Z^{*})\right)^{-1} \frac{\partial \varphi(Z^{*}, X, \mathcal{G})}{\partial (\cdot)}, 
\end{equation}
where $J_{\varphi}(Z^{*}) = \frac{\partial \varphi(Z^{*}, X, \mathcal{G})}{\partial Z^{*}}$.

Combining Equation \eqref{eq:Z*_derivative} and \eqref{eq:gradients_w_z*}, we can have the following:
\begin{equation}
    \frac{\partial \ell}{\partial (\cdot)} = \frac{\partial \ell}{\partial Z^{*}} \left(I - J_{\varphi}(Z^{*}) \right)^{-1} \frac{\partial \varphi(Z^{*},X,\mathcal{G})}{\partial (\cdot)}.
\end{equation}

\section{Inefficiency of using higher $\gamma$}
\label{appendix: inefficiency and instability}
Using a large contraction factor $\gamma$ usually make the process of finding the fixed-point more instable and difficult and it requires more iterations to find the fixed-point solution, which compromises the efficiency. 
We conduct the empirical experiments to verify this. We use EIGNN \citep{EIGNN} with iterative solvers as the model on the chain dataset (as described in Sec \ref{sec:range_analysis}).
Table \ref{tab:higher_gamma} demonstrates the training time of using different values of $\gamma$.
We can see that when using $\gamma=0.9$ can cost 2x training time compared with using $\gamma=0.8$. 
Smaller $\gamma$ empirically causes faster convergence to get the fixed-point solutions. 
\begin{table}[h]
    \centering
    \begin{tabular}{c|cccc}
         \toprule
         $\gamma$ & 0.6 & 0.8 & 0.9 & 0.95   \\
         \midrule
          Time per epoch & 0.86s  & 1.65s & 3.71s & 3.95s \\
         Total time & 4596s & 8267s & 18580s & 19916s \\
         \bottomrule
    \end{tabular}
    \caption{Training time with different $\gamma$ used in the iterative method.}
    \label{tab:higher_gamma}
\end{table}

Besides EIGNN, IGNN also faces instability in the training if it uses a large contraction factor. 
Different with Equation \eqref{eq:Implicit_propagation}, IGNN projects the weight matrix $W$ with a contraction factor $\kappa$ onto a convex constraint set to ensure the convergence of iterative mapping. 
In their official implementation repo \footnote{https://github.com/SwiftieH/IGNN/issues/3}, they mention that too large $\kappa$ may cause the non-convergence of the equilibrium equation, which leads significant performance degradation. 
In the training log they provided \footnote{https://github.com/SwiftieH/IGNN/files/7052441/PPI\_output\_log.txt},
with $\kappa=0.98$, we can see that the loss suddenly jump to more than 2000 from around 0.019 and the accuracy degrades from 0.96 to 0.39.
This verifies again that a large contraction factor may cause instability during the training, although EIGNN and IGNN use different ways for contraction. 
\section{More on Experiments}
\label{appendix: details of experiments}
\begin{figure}
    \centering
    \begin{minipage}{.48\textwidth}
        \includegraphics[width=1\textwidth]{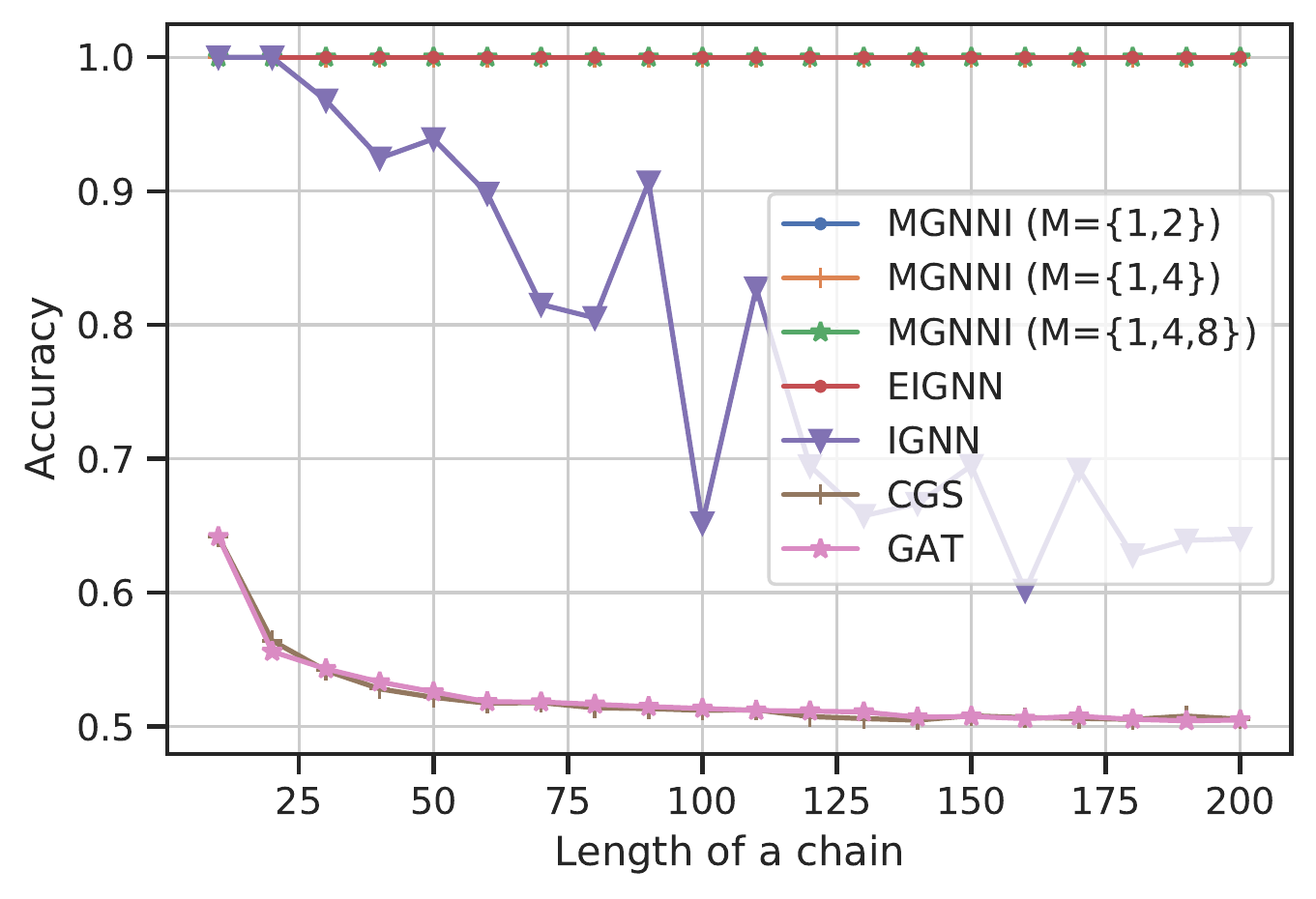}
        \caption{Averaged accuracies on Chains dataset: compare \ourmodel to
        implicit and explicit GNNs.}
        \label{fig:chain_bincls}
    \end{minipage} \quad
    \begin{minipage}{.48\textwidth}
        \includegraphics[width=1\textwidth]{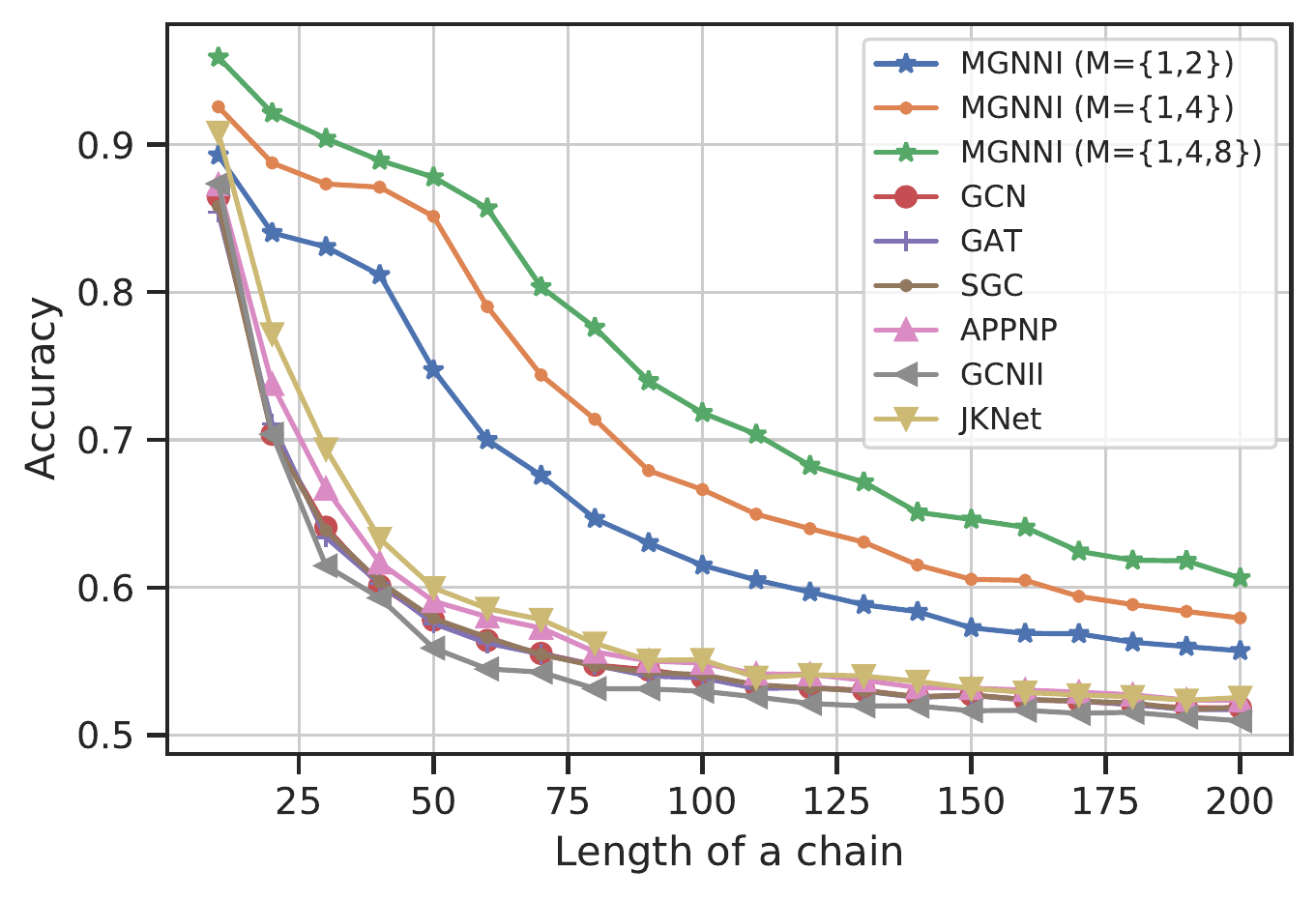}
        \caption{Averaged accuracies on color-counting dataset: compare \ourmodel with explicit GNNs.}
        \label{fig:counting_explicitGNNs}
    \end{minipage}
    % \includegraphics{Figs/binary-counting_chains_explicitGNNs.pdf}
    % \caption{Averaged accuracies on color-counting dataset: compare \ourmodel with explicit GNNs.}
    % \label{fig:counting_explicitGNNs}
\end{figure}
\subsection{Synthetic experiments}
\label{appendix:synthetic_exp}
\paragraph{Chains dataset}
The synthetic chain dataset is used in \citet{IGNN} and \citet{EIGNN} to evaluate the ability of models to capture distant information. 
Chain dataset contains several chains directed from one end to the other with length $l$. 
The label information is only encoded as the node features in the starting end node of each chain.
The nodes on the same chains share the same class label. 
The task is to classify nodes into different classes, which requires simply passing information from one end to the other end of a chain graph. 
We consider binary classification, 20 chains for each class, and $l$ nodes in each chain. 
Then, the chain dataset has $40\times l$ nodes. 
We randomly split the dataset for train/val/test as 5\%/10\%/85\%. 

In Figure \ref{fig:varing_gamma} of Section \ref{sec:range_analysis}, the starting end node is regarded as node $p$. 
We perturb the node features of starting node $p$ by masking the features to all zeros.
After that, we measure the L2 norm of the change in node $q$'s equilibrium as increasing the distance between node $q$ and $p$ to support our analysis.

Here, as in \citet{EIGNN} and \citet{IGNN}, we also provide the averaged results over 20 runs on the Chains dataset for comparison between \ourmodel and other representative baselines in Figure \ref{fig:chain_bincls}. 
As we can see, \ourmodel with different scale combinations can all achieve 100\% accuracy as EIGNN, which indicates the superiority of capturing long-range dependencies. 
In contrast, IGNN has decreasing accuracies as the chain length increases. 
CGS has similar performance with GAT, which demonstrates the deficiency in capturing long-range information. 
We conjecture the reason is that CGS uses a single layer attention variant of graph network (GN) \citep{battaglia2018relational} as the input transformation which is a finite GNN as GAT. 
For simplicity, we omit more results of other explicit GNNs as those results can provided by \citet{EIGNN} in their Figure 1. Note that those explicit GNNs all generally perform worse than IGNN and EIGNN.

% \TODO{explain non-zero acc of finite GNNs}
\paragraph{More results on color-counting dataset}
Besides the comparison between \ourmodel and other implicit GNNs, we also compare \ourmodel with explicit GNN models. 
Figure \ref{fig:counting_explicitGNNs} shows that \ourmodel with different scales consistently outperform all explicit GNNs, which confirms that \ourmodel as an implicit GNN model has the better ability to capture long-range dependencies compared with explicit GNNs. 
Explicit GNN models still can achieve more than 85\% accuracy when the chain length is 10. It is because that the test set are randomly sampled where some test nodes may be placed nearby the starting end node. 
However, the performance of explicit GNNs drops quickly as the chain length increases. 

\subsection{Node classification on real-world datasets}
\label{appendix:node_cls_real_world}
\paragraph{Dataset descriptions}
We first use 5 heterophilic graph datasets as in \citet{Pei2020Geom-GCN} to evaluate the capability of capturing long-range dependencies:
\begin{itemize}
    \item \textbf{Chameleon and Squirrel}: these graphs are originally collected by \citet{musae}, using the web pages in WikiPedia of the corresponding topic. Nodes represent web pages and edges are hyper-links from a web page to another. The class labels are generated by \citet{Pei2020Geom-GCN}. There are 5 categories indicating the amount of the average monthly traffic of web pages. 
    \item \textbf{Cornell, Texas, and Wisconsin}: these datasets contain the web-page graphs of the corresponding universities. Label classes indicate the category of web pages, where 5 classes are considered, i.e., student, faculty, course, project, and staff. These three datasets are collected by the CMU WebKB project \footnote{\url{http://www.cs.cmu.edu/~webkb/}}. The preprocessed version generated by \citet{Pei2020Geom-GCN} is used in our experiments. 
\end{itemize}
To evaluate the model capacity on multi-label multi-graph inductive setting, we conduct the experiment on Protein-Protein Interaction (PPI) dataset. 
The dataset is originally collected from the Molecular Signatures Database \citep{subramanian2005gene} by \citet{Graphsage}. 
PPI dataset has 24 graphs, where each graph represents a different human tissue. 
Each graph has nodes representing proteins and edges indicating interactions between proteins. 
Each node can have maximum 121 labels which represents gene ontology sets. 
We use the same data splits as in \citet{Graphsage}, i.e., 20 graphs for training, two graphs for testing, and two other graphs used for validation. 

\paragraph{Experimental settings}
For heterophilic graphs, we compare \ourmodel with 3 implicit GNNs (i.e., IGNN \citep{IGNN}, EIGNN \citep{EIGNN}, and CGS \citep{CGS}) and 8 explicit GNNs (i.e., Geom-GCN \citep{Pei2020Geom-GCN}, SGC \citep{SGC}, GCN \citep{semi_GCN}, GAT \citep{GAT}, APPNP \citep{APPNP}, JKNet \citep{JKNet}, GCNII \citep{GCNII}, and H2GCN \citep{H2GCN}). 
As we follow the exact same setting as in \citep{EIGNN}, we reuse their results of baselines, except CGS. For CGS and \ourmodel, we conduct the experiments with 20 different runs and report the averaged accuracies with standard deviation. 

For network architectures used in \ourmodel on heterophilic graphs, we use two-layer MLP followed by the ReLU function as input features transformation $f(\cdot)$ and a linear map as the output transformation function (i.e., $f_o(X) = WX$). The hyperparameter search space is set as follows: multiscale set $M$ \{\{1,2\}, \{1,3\}, \{1,2,3\}\}, weight decay \{5e-6, 5e-4\}, learning rate \{0.01, 0.05, 0.1, 0.5\}.
We use $\gamma = 0.8$ for all scale modules and 0.5 as the dropout rate. The Adam optimizer \citep{kingma:adam} is used for optimization. 
For CGS, we use the suggested network architectures (i.e., a single layer attention variant of graph network (GN) \citep{battaglia2018relational} and the same number of hidden neurons as in CGS paper \citep{CGS}. For other hyperparameter tuning, we optimize over learning rate \{0.001, 0.005, 0.01, 0.05\}, weight decay \{5e-6, 5e-4\}, and $\gamma$ \{0.5, 0.8\}.

For PPI datasets, we use a 4-layer MLP directly after the multiscale propagation, while IGNN applies 4 MLPs between four consecutive IGNN layers. We set \{1,2\} as the multiple scales in our propagation module, and use 0.001 as the learning rate. No dropout is used.

\subsection{Graph classification on real-world datasets}
\label{appendix:graph_cls}
\paragraph{Dataset descriptions}
We conduct experiments on 4 bioinformatics datasets (MUTAG, PTC, PROTEINS, NCI1) and 2 social-network datasets (IMDB-Binary and IMDB-Multi), following identical settings as in \citep{GIN, IGNN}. 
MUTAG is a dataset having 188 graphs representing mutagenic aromatic and heteroaromatic nitro compounds with 7 discrete labels. 
PTC is a dataset of 344 chemical compounds reporting the carcinogenicity for male and female rats and it has 19 discrete labels.
PROTEINS is a dataset where nodes are secondary structure elements (SSEs) and an edge between two nodes indicates they are neighbors in the amino-acid sequence or in 3D space. It has 3 discrete labels, representing sheet, helix or turn.
NCI1 is a subset of balanced datasets of chemical compounds screened for ability to suppress or inhibit the growth of a panel of human tumor cell lines with 37 discrete labels and it is made publicly available by the National Cancer Institute (NCI). 

IMDB-Binary and IMDB-Multi are social-network datasets, indicating movie collaborations. Each graph contains a ego-graph for each actor/actress, where nodes represent actors/actresses and edges connect two actors/actresses if they appear in the same movie. 
Labels are pre-specified genres of movies and each graph has the label corresponding to its genre. The task requires models to classify the genre. 
\paragraph{Experimental settings}
We compare \ourmodel with several representative baselines, including several explicit GNNs: Graph Convolution Network (GCN) \citep{semi_GCN}, Deep Graph Convolutional Neural Network (DGCNN) \citep{zhang2018end}, Fast and Deep Graph Neural Network (FDGNN) \citep{gallicchio2020fast}, and Graph Isomorphism Network (GIN) \citep{GIN}, and two other implicit GNNs: Implicit Graph Neural Network (IGNN) \citep{IGNN} and Convergent Graph Solver (CGS) \citep{CGS}. 
% \subsection{Experimental setting}

Since we follow the same experimental settings as in \citep{IGNN, GIN}, we reuse the results of baselines from \citep{IGNN, CGS}, except EIGNN \citep{EIGNN}.
For the network architectures used in \ourmodel for graph classification, we use three-layer MLP followed the ReLU function as input feature transformation $f(\cdot)$. 
After the multiscale propagation with attention mechanism, we use the sum-pooling aggregator to obtain the graph representations. A linear map is used as the output transformation function (i.e., $f_o(X_g) = WX_g$). 
We set the number of hidden state as 32, $y = 0.8$ for all scale modules. The search space for the other hyperparameters is set as follows: multiscale set $M$ \{\{1,2\}, \{1,3\}\}, weight decay \{0, 5e-6\}, and learning rate \{0.001, 0.01\}. The Adam optimizer \citep{kingma:adam} is used for optimization. 
For EIGNN, after their implicit layer, we use a 3-layer MLP with ReLU activation followed the sum-pooling layer to obtain the graph representations. A 2-layer MLP is used for generating the final predictions. $\gamma$ is set to 0.8 and other hyperparameters (learning rate and weight decay) are tuned over the same search space.

\subsection{Efficiency Comparison}
\label{app:efficiency_comparsion}
\begin{wraptable}{r}{0.4\textwidth}
    \centering
    \caption{Training time per epoch on PPI.}
    \label{tab:PPI_efficiency}
    \begin{tabular}{@{}lcc@{}}
    \toprule
    Method                 & Train Time & Pre-process     \\ \midrule
    IGNN                 & 32.7s  & N.A.        \\ 
    EIGNN               & 2.3s & 45s\\ \midrule
    \ourmodel & 2.6s & N.A. \\
    \bottomrule
    \end{tabular}
\end{wraptable}
Here, we provide the efficiency comparison among IGNN, EIGNN, and \ourmodel on PPI dataset. 
Table \ref{tab:PPI_efficiency} demonstrates the training time per epoch and the total pre-processing time of different models. 
We can see that IGNN requires around 30s for training an epoch, while EIGNN and \ourmodel requires only around 2s for an epoch. 
The reason of inefficiency in IGNN is that IGNN requires 4 implicit layers sequentially stacked, which means that every iterative solver needs to wait the fixed-point solution provided by the previous iterative solver for solving its own solution. 
In contrast, \ourmodel have parallel equilibrium layers with different scales, where each equilibrium layer can get the fixed-point solution simultaneously.
Thus, the training time per epoch of \ourmodel is similar with that of EIGNN which only has one implicit layer. 
Additionally, \ourmodel does not require pre-processing time to conduct eigendecomposition of the adjacency $S$ as EIGNN which may be costly for large graphs.

\begin{wraptable}{r}{0.45\textwidth}
    \centering
    \vspace{-0.7cm}
    \caption{Performance of different scales.}
     \label{tab:removing_attention}
    \begin{tabular}{@{}cccc@{}}
        \toprule
         Scales & PPI & Chameleon & Texas \\
         \midrule
         (wo/ att) \{1,2\} & 98.35 & 61.46 & 81.35\\
         (w/ att) \{1,2\} & 94.62 & 58.24 & 82.97 \\
         \midrule
         (wo/ att) \{1,2\} & 98.67 & 63.93 & 83.24\\
         (w/ att) \{1,2\} & 98.74 & 63.75 & 84.86 \\ 
        \bottomrule
    \end{tabular}
\end{wraptable}
\subsection{The effect of the attention mechanism}
\label{appendix: attention mechanism effect}
To quantitatively investigate the effect of the attention mechanism in MGNNI, we conduct additional experiments by removing the attention mechanism and instead use average pooling for fusing information from multiple scales. 
The experimental results are provided in Table \ref{tab:removing_attention}. We can see that, if we replace the attention mechanism with average pooling, the performance would drop. It verifies the effectiveness of our attention mechanism, which is also demonstrated in Figure \ref{fig: att_vals} and its corresponding explanations in Section \ref{sec: ablation study}. 

\end{document}